\documentclass[lettersize,journal]{IEEEtran}
\usepackage{amsmath,amsfonts}
\usepackage{algorithmic}
\usepackage{algorithm}
\usepackage{array}
\usepackage[caption=false,font=normalsize]{subfig}
\usepackage{textcomp}
\usepackage{stfloats}
\usepackage{url}
\usepackage{verbatim}
\usepackage{graphicx}
\usepackage{cite}
\usepackage{bbm}

\usepackage{algorithm}
\usepackage{algorithmic}

\usepackage{xspace}
\newcommand{\sexyname}{DSM-NAS\xspace}
\newcommand{\sexynameplus}{DSM-NAS+\xspace}

\usepackage{color}
\usepackage{xcolor}

\usepackage{multirow}
\usepackage{makecell}
\usepackage{bbding}

\usepackage{graphicx}
\usepackage{caption}

\usepackage{lineno}
\usepackage{hyperref}
\hypersetup{colorlinks=true,urlcolor=blue}

\usepackage{booktabs} 
\newcommand{\topline}{\toprule [0.1em]}
\newcommand{\midline}{\midrule [0.05em]}
\newcommand{\bottomline}{\bottomrule [0.1em]}

\def\mE{{\mathcal E}}

\def\mG{{\mathcal G}}

\def\mV{{\mathcal V}}

\DeclareMathAlphabet\mathbfcal{OMS}{cmsy}{b}{n}

\def\0{{\bf 0}}
\def\1{{\bf 1}}

\def\mmE{{\mathbb E}}

\def\eg{\emph{e.g.}} 
\def\ie{\emph{i.e.}} 
 
 \def\vs{\emph{vs.}}
\def\wrt{{w.r.t.~}}

\usepackage{amsmath}

\begin{document}

\title{Automated Dominative Subspace Mining for \\ Efficient Neural Architecture Search}

\author{
Yaofo Chen, Yong Guo, Daihai Liao, Fanbing Lv,  Hengjie Song, \\ James Tin-Yau Kwok, \textit{IEEE Fellow}, and Mingkui Tan, \textit{IEEE Member}
\thanks{
	Yaofo Chen, Hongjie Song and Mingkui Tan
	are with the School of Software Engineering,
	South China University of Technology,
	Guangzhou 510641,
	China
	(e-mail:
	chenyaofo@gmail.com;
	sehjsong@scut.edu.cn;
	mingkuitan@scut.edu.cn)
}
\thanks{Yong Guo is with the Max Planck Institute for Informatics, Max Planck Institute, Saarbrücken 66123, Germany. (e-mail: guoyongcs@gmail.com).}
\thanks{
	Daihai Liao and Fanbing Lv are with Changsha Hisense Intelligent System Research Institute Co., Ltd, Changsha 410006, China
	(e-mail:
	liaodaihai@hisense.com;
	lvfanbing@hisense.com)
}
\thanks{
    James Tin-Yau Kwok is with the Department of Computer Science and Engineering, the Hong Kong University of Science and Technology, Hong
Kong 999077, China
	(e-mail:  jamesk@cse.ust.hk)
}
\thanks{
This work was partially supported by National Natural Science Foundation of China (NSFC) 62072190, the Research Grants Council of the Hong Kong Special Administrative Region, China (Project No. 16202523 and HKU C7004-22G), the Major Key Project of Peng Cheng Laboratory (PCL) PCL2023A08, the Young Scholar Project of Pazhou Lab (No.PZL2021KF0021), TCL Science and Technology Innovation Fund.
}
\thanks{Yaofo Chen and Yong Guo contributed equally to this paper.}
\thanks{Mingkui Tan is the corresponding author.}
}

\markboth{Journal of \LaTeX\ Class Files,~Vol.~14, No.~8, August~2021}%
{Shell \MakeLowercase{\textit{et al.}}: A Sample Article Using IEEEtran.cls for IEEE Journals}


\maketitle

\begin{abstract}
Neural Architecture Search (NAS) aims to automatically find effective architectures within a predefined search space.
However, the search space is often extremely large.
As a result, directly searching in such a large search space is non-trivial and also very time-consuming.
To address the above issues, in each search step, we seek to limit the search space to a small but effective subspace to boost both the search performance and search efficiency.
To this end, we propose a novel Neural Architecture Search method via Dominative Subspace Mining (\sexyname) that finds promising architectures in automatically mined subspaces.
Specifically, we first perform a global search, \ie, dominative subspace mining, to find a good subspace from a set of candidates.
Then, we perform a local search within the mined subspace to find effective architectures.
More critically, we further boost search performance by taking well-designed/searched architectures to initialize candidate subspaces.
Experimental results demonstrate that \sexyname not only reduces the search cost but also discovers better architectures than state-of-the-art methods in various benchmark search spaces.
\end{abstract}


\section{Introduction}
\IEEEPARstart{D}{eep} neural networks (DNNs) have been the workhorse of many challenging tasks, including image classification~\cite{resnet,alexey2021vit,liu2021Swin}, action recognition~\cite{duan2020omni,feichtenhofer2019slowfast} and natural language processing~\cite{radford2018improving,brown2020language,song2021richly,hu2022expansion}.
The success of DNNs is largely attributed to the innovation of effective neural architectures.
However, designing effective architectures often greatly depends on expert knowledge and human efforts.
Thus, it is non-trivial to design architectures to satisfy the requirements manually.
To address this, neural architecture search (NAS)~\cite{zoph2016neural} is developed to automate the process of the architecture design.

Existing NAS methods search for effective architectures in a predefined search space~\cite{zoph2016neural,liu2018darts,tan2019mnasnet}.
To cover as many good architectures as possible, the search space is often designed to be extremely large (\eg, $\small{\sim}10^{12}$ in ENAS~\cite{pham2018efficient} and $\small{\sim}10^{19}$ in OFA~\cite{Cai2020Once}).
Directly searching in such a large space is very difficult and time-consuming in practice~\cite{guo2020breaking}.
Specifically, to explore the large search space, we have to sample and evaluate plenty of architectures, which is very computationally expensive and time-consuming.
Moreover, we can only access a small proportion of architectures in the search space due to the limitation of the computational resources in practice.
In other words, regarding a very large search space, we can only obtain limited information to guide the architecture search.

\begin{figure*}[t]
\centering
\includegraphics[width=1\linewidth]{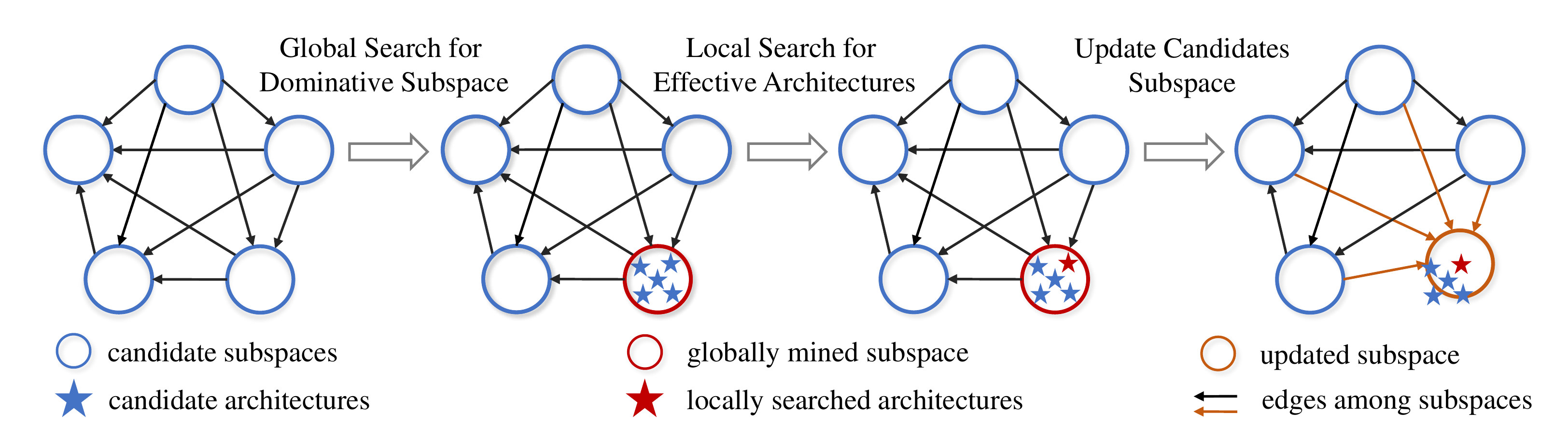}
\caption{An illustration of the search process.
We find promising architectures in a two-step search manner:
1) we perform global search to mine/find a dominative subspace from a set of candidates;
2) we move the focus to the subspace and conduct a local search for effective architectures within it.
Then, we update the candidate subspace with the better searched architecture.
}
\label{fig:graph_space}
\end{figure*}

To overcome the above difficulties brought by the large search space, PNAS~\cite{liu2018progressive} and CNAS~\cite{guo2020breaking} propose to start from a very small search space to perform an architecture search and then gradually enlarge the search space by adding nodes or operations.
Recently, AlphaX~\cite{wang2019alphax} partitions the search space into existing good subspaces and unexplored subspaces and adopts the Monte Carlo Tree Search (MCTS) method to encourage exploring the good ones.
However, these methods suffer from two limitations.
\textbf{First}, these methods still find architectures from a very large space at each search step, which may not only result in unnecessary explorations but also affect the search results.
\textbf{Second}, the small search spaces partitioned by these methods are fixed during the search phase, which may not be optimal to find good architectures.
Thus, how to find/design a small but effective search space that covers as many good architectures as possible is an important problem.

To achieve this goal, an underlying hypothesis is that the neighborhood around an effective architecture is usually a good subspace for further exploration.
In this case, once we find the small but effective search space around a promising architecture, it is more likely to find a better architecture within the subspace rather than the entire search space.
We empirically verify this hypothesis that similar architectures tend to have close performance (See the results of Figure~\ref{fig:acc_vs_dist}, as well as the observations in~\cite{ying2019bench,guo2020breaking}).
Inspired by this, instead of searching in the whole search space, we seek to limit the search space to a reduced one in each search step by recognizing and mining a small but effective subspace.
In this sense, we may boost the search performance and search efficiency as well since we only focus on a mined effective subspace, in which it is easier to find good architectures than directly exploring the whole search space.

In this paper, we propose an efficient Neural Architecture Search method via Dominative Subspace Mining (called \sexyname). 
The key idea is to find/mine a small but effective/dominative subspace from the whole search space in each step of the architecture search.
To this end, we first construct a set of candidate subspaces and then build a subspace graph from them, with edges denoting the relationships/information among different subspaces.
As shown in Figure~\ref{fig:graph_space}, 
we first perform a \textit{global search} to automatically mine a dominative subspace from the subspace graph.
Then, we focus on the mined subspace and conduct a \textit{local search} to obtain the resultant architectures.
It is worth noting that once we find a better architecture, we also update the subspace graph accordingly.
In this way, it becomes possible to gradually find better architectures during the search process.
Moreover, we are able to further boost search performance of the proposed \sexyname by taking existing well-designed architectures (\eg, OFA architecture~\cite{Cai2020Once}) to construct candidate subspaces.
Extensive experiments in two benchmark search spaces demonstrate the effectiveness of the proposed method.

Our contributions are summarized as follows:
\begin{itemize}
    \item 
    Instead of searching in the entire search space, we seek to find/mine a small but effective/dominant subspace for each step in the architecture search.
    With the help of the mined subspaces, we are able to improve search performance and efficiency.
    
    \item We propose a novel Dominative Subspace Mining algorithm to enhance the performance of neural architecture search. 
    Specifically, we first perform a global search to find dominative subspaces and then perform a local search to obtain the resultant architectures.
    
    \item Extensive experiments demonstrate the superiority of the proposed method over existing NAS methods. More importantly, the searched subspaces also exhibit promising transferability to new datasets.
\end{itemize}

\section{Related Work}

\subsection{Neural Architecture Search}

In recent years, NAS~\cite{wu2022iddnet,mei2023automatic,guo2023latency} has drawn great attention for effective architecture design.
Traditional reinforcement learning-based NAS methods~\cite{baker2016designing,guo2021towards,chen2021contrastive,liu2022data,yuzuguler2022u} directly maximizes the expectation of the performance of searched architectures sampled from the whole search space. In contrast, our \sexyname searches in a small and effective subspace instead of the whole space by employing both global search and local search policies. In this sense, \sexyname alleviates the difficulties due to a large search space.
Gradient-based NAS methods~\cite{liu2018darts,chen2021drnas,Xie_2022_CVPR,Xiao_2022_CVPR,Zhang_2023_CVPR,han2023differentiable} find effective architectures by relaxing the search space to continuous-valued and optimizing by gradient descent. They introduce differentiable architecture parameters, in which the network weights and architecture parameters are alternately optimized. Our proposed method focuses on searching in small and effective subspaces instead of the entire large space. This is a significant difference from gradient-based methods, which typically operate in a continuous relaxation of the whole search space.

Besides, Evolutionary-based NAS methods~\cite{lu2020neural,chen2021oneshot,chen2021autoformer,dai2021fbnetv3,Pan_2022_CVPR} find promising architectures through crossover and mutation in the neighborhood/subspace of the population.
Our \sexyname differs from these methods in three significant ways.
First, they usually perform crossover and mutation randomly. 
Our \sexyname searches architectures in the mined subspace with a learned policy, which results in higher search efficiency.
Second, they usually greedily select the best architecture from the population for reproduction. In contrast, our \sexyname finds a promising subspace that has the potential to achieve the largest performance improvement, which helps to encourage the exploration ability (See Figure~\ref{fig:comparisons_with_rewards}).
Third, during mutation, they treat elements in the population independently to find a subspace. 
Our method builds a subspace graph to exploit the relationship among different subspaces to enhance the search performance (See  Figure~\ref{fig:comparisons_with_subgraph}).

However, traditional NAS methods~\cite{baker2016designing,real2019regularized} often require great computational resources and thus result in unaffordable time costs.
A lot of efforts have been made to improve the search efficiency of the search process. They mainly focus on improving the efficiency of architecture performance estimation.
Specifically, weight-sharing-based NAS methods~\cite{pham2018efficient,Cai2020Once} estimate the performance of candidate architectures without the need to train each one from scratch.
Instead, they train a supernet that encompasses all possible architectures under consideration. While estimating architectures, their weights are inherited from the trained supernet, which greatly lowers the computational cost of architecture evaluation.
EcoNAS~\cite{zhou2020econas} observes that most existing proxies exhibit different behaviors in maintaining rank consistency among architectures. Based on this, it designs a proxy training strategy that is potentially more accurate in evaluating architectures on small proxy datasets.

Besides, Zero-cost proxy methods (such as Grad-norm~\cite{mohamed2021zero}, SNIP~\cite{lee2019snip}, GraSP~\cite{wang2020picking} and Synflow~\cite{tanaka2020pruning}) reduce the computational cost by summing up the saliency value of the model weights. ZiCO~\cite{li2023zico} devises a new architecture estimation strategy by calculating \textit{Zero-shot inverse Coefficient of Variation} scores with a single forward/backward propagation. These methods are very efficient since they just require a single forward/backward propagation while evaluating an architecture. 
This is in stark contrast to traditional methods that may require extensive training and multiple iterations to obtain the performance of an architecture.
Unlike these methods which focus on rapidly evaluating architectures, our method achieves high efficiency from a different perspective. To be specific, we reduce the number of architecture evaluations required to identify promising architectures. To achieve this, our \sexyname focuses on small and effective subspaces that are more likely to contain high-performing architectures. By narrowing down the search to these subspaces, the number of required architecture evaluations decreases. This not only accelerates the search process but also ensures that the computational resources are expended on evaluating architectures that have a higher likelihood of success.

\subsection{Search Space Design of NAS Methods}

NAS methods often find promising architectures in a predefined large search space, such as NASNet~\cite{zoph2018learning}, DARTS~\cite{liu2018darts} and MobileNet-like~\cite{tan2019mnasnet} search spaces.
Most existing methods directly perform search in these large search spaces, which may not only result in inefficient sampling but also hamper the search performance.
To address this issue, local search methods~\cite{white2020exploring} search in an iterative manner. They visit architectures in the neighborhood of an architecture and update it with the best-found architecture. 
This is computationally expensive since it requires plenty of architecture evaluations.
Recent works, such as PNAS~\cite{liu2018progressive} and CNAS~\cite{guo2020breaking}, implement a progressive space-growing search strategy. 
They start from a compact search space to mitigate the challenges posed by a large search space, and subsequently expand the search space by incorporating additional nodes or operations. However, these methods suffer from two limitations. First, in the later stages of the search, these methods still try to identify architectures within an exceedingly large space, potentially leading to insufficient explorations. Secondly, the small search spaces delineated by these methods are arbitrarily created through the addition of nodes or operations and remain static throughout the search process.
This may not be optimal for discovering high-performing architectures. In contrast, our \sexyname consistently searches in small yet potent subspaces throughout the entire search process and dynamically identifies improving and dominant subspaces.

AlphaX~\cite{wang2019alphax} and LaNAS\cite{wang2022sample} build a Monte Carlo Search Tree to partition the search space into different subspaces according to their performance and encourage exploration of the promising subspaces. Nonetheless, note that the partitioning process is bound by a predetermined tree structure, which imposes certain limitations on the method.
During each step of the search, the algorithm is restricted to partitioning based on the available candidate operations.
In contrast, our \sexyname adopts a more flexible and unconstrained approach to navigating the search space. It focuses on identifying and mining the dominant subspace, which is determined solely based on the central architecture of that subspace. It is important to note that this central architecture is not fixed, but instead can be any configuration in the entire search space, thus providing a broader exploration.

Besides, Few-shot NAS~\cite{zhao2021few} adopts multiple sub-supernets to encompass different regions (\ie, subspaces) of the search space, aiming for a precise evaluation of architectural performance. These subspaces are different from the entire search space at the beginning and remain constant in subsequent iterations.
In contrast, our \sexyname differs from these methods in two significant ways: 1) \sexyname tries to mitigate the difficulties incurred by a large search space rather than enhancing architectural performance estimation; 2) \sexyname adopts a dynamic approach, actively searching for the dominant subspaces, instead of relying on static, predetermined subspaces.
RegNet~\cite{ilija2020designing} designs an initial search space that is very large, and then subsequently narrows it down to a more refined and compact search space through empirical analysis of the architectural behaviors. It is imperative to highlight that the entire pipeline operates manually. In contrast, our \sexyname automatically explores dominative search spaces and identifies promising architectures without the need for human intervention, thanks to our innovative global and local search strategies.

\begin{figure*}[t]
\centering
\includegraphics[width=.95\linewidth]{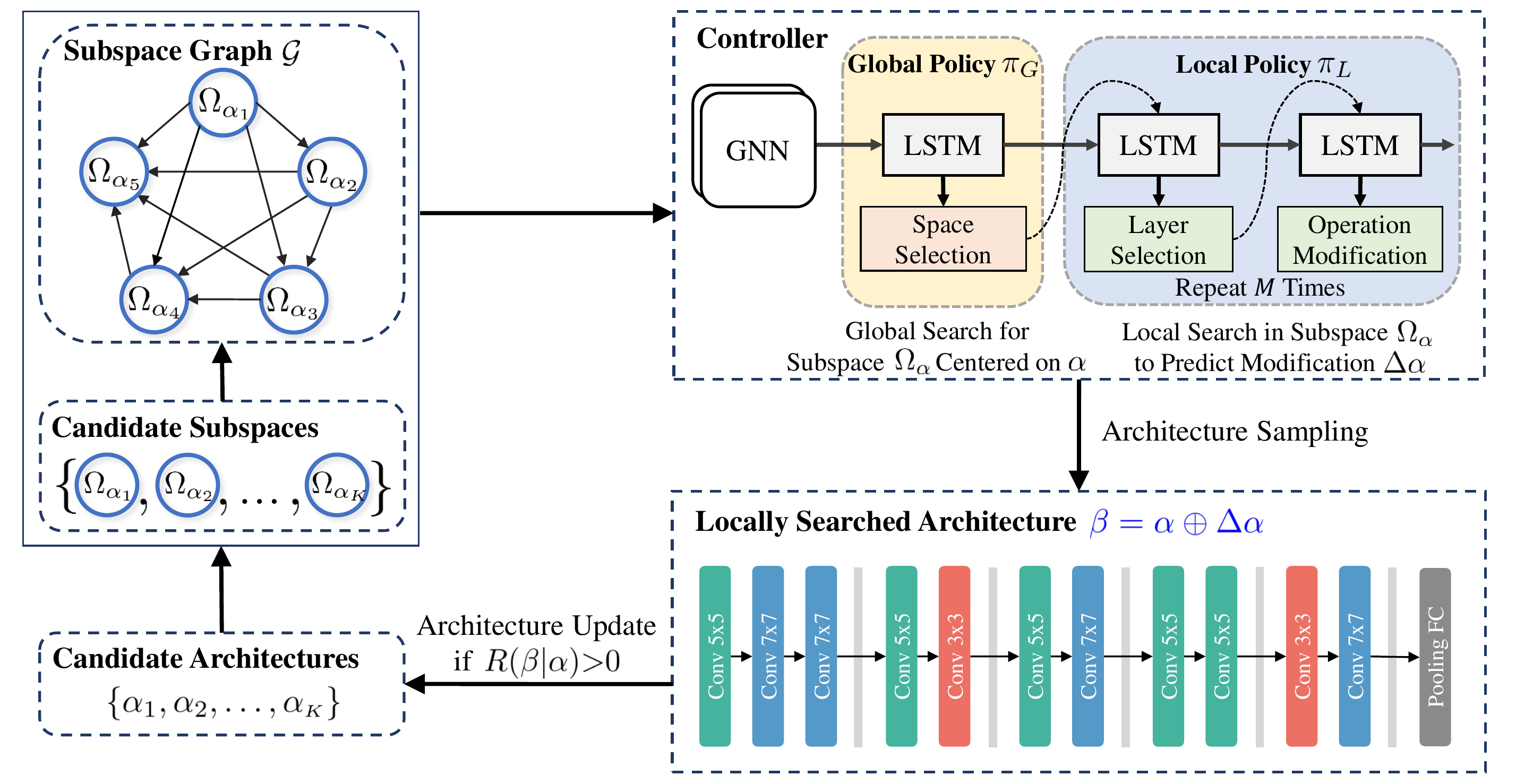}
\caption{An overview of the proposed \sexyname. We build a set of subspaces $\{\Omega_{\alpha_i}\}_{i=1}^K$ centered on randomly sampled candidate architectures $\{\alpha_i\}_{i=1}^K$ and construct a subspace graph $\mG$ to model the relationships among these subspaces.
By taking $\mG$ as the input, the controller first mines/finds a dominative subspace $\Omega_\alpha \sim \pi_{G}(\cdot | \mG; \theta_{G})$ via global search and then predicts an architecture modification $\Delta \alpha \sim \pi_{L}(\cdot | \Omega_{\alpha}; \theta_{L})$ via local search.
Next, we update the candidate architecture $\alpha$ with the resultant architecture $\beta=\alpha \oplus \Delta \alpha$ if $\beta$ has better performance than $\alpha$ (\ie, $R(\beta | \alpha) > 0$).
}
\label{fig:overview}
\end{figure*}

\section{Architecture Search via Subspace Mining}

In this paper, we propose a neural architecture search method with Dominative Subspace Mining (\sexyname) to boost both the search performance and efficiency of NAS.
In Section~\ref{sec:overall}, we first discuss the motivation and provide an overview of \sexyname.
Then, we describe the details of the two key steps in our method, \ie, the global search and local search in Sections~\ref{sec:global_search} and ~\ref{sec:local_search}, respectively.
In Table~\ref{tab:notations}, we present some frequently used notations.

\begin{table}[t]
\centering
\caption{Summary of frequently used notations.}
\resizebox{\linewidth}{!}{
\begin{tabular}{c|l}
    \topline
    Notation &  Description  \\
    \midline
    $\Omega$ & search space \\
    $\alpha$ & architecture \\
    $\Omega_{\alpha}$ & search subspace centered on an architecture $\alpha$ \\
    $R(\alpha, w_{\alpha})$ & performance metric of an architecture $\alpha$ \\
    $R(\beta | \alpha)$ & performance improvement between $\beta$ and $\alpha$ \\
    $\pi_{G}$ & global search policy \\
    $\pi_{L}$ & local search policy \\
    $\mathcal{G}$ & subspace graph \\
    $K$ & number of candidate subspaces in $\mathcal{G}$ \\
    $M$ & local search distance \\
    $\Delta\alpha$ & architecture modification \\
    $\oplus$ & combination operation \\
    \bottomline
\end{tabular}
}
\label{tab:notations}
\end{table}

\subsection{Motivation and Method Overview}
\label{sec:overall}
Existing NAS methods often consider an extremely large search space $\Omega$ to find good architectures~\cite{pham2018efficient,liu2018darts, Cai2020Once}. 
However, directly performing architecture search in such a large search space is non-trivial and often very expensive.
Instead of using the whole search space, it should be possible to limit each search step within a reduced search space to boost search performance and search efficiency.
To achieve this goal, an underlying hypothesis is that the neighborhood around an effective architecture is usually a promising/dominative subspace for further exploration to find better ones~\cite{ying2019bench,guo2020breaking}.
In this case, we are more likely to find effective architectures in the dominative subspace than the whole space under the same search budget.

Inspired by this, we propose a new search algorithm by mining/identifying effective subspaces, in which 
it is easier to find good architectures than directly exploring the whole space.
To this end, we define a subspace $\Omega_\alpha$ based on a center architecture $\alpha$ within it (See more details in Section~\ref{sec:global_search}).
As shown in Figure~\ref{fig:overview} and Algorithm~\ref{alg:training}, we first learn a global search policy $\pi_G$ to automatically search for a dominative subspace $\Omega_\alpha$ based on a center architecture $\alpha$.
Then, we further learn a local search policy $\pi_L$ to produce the resultant architectures in the subspace.
Specifically, the global search policy $\pi_G$ takes a set of candidate subspaces $\{\Omega_{\alpha_i}\}_{i=1}^{K}$ (as well as their relationship) as inputs and mines/finds a dominative subspace $\Omega_{\alpha}$.
Based on the mined/searched subspace, the local policy $\pi_L$ further generates a modification $\Delta\alpha$ (\ie, modifying some operations of some layers in $\alpha$) to explore the subspace.
Finally, we combine $\alpha$ and $\Delta \alpha$ to obtain the resultant architecture by
\begin{equation}
    \beta {=} \alpha {\oplus} \Delta \alpha.
\end{equation}
Here, $\oplus$ denotes the combination operation, as will be described in Section~\ref{sec:local_search}.
Note that $\Delta \alpha$ is devised to constrain the architecture after modifications still in the subspace.

Note that if we directly maximize the performance of the resultant architecture $\beta$, the search algorithm may always select the same subspace with the best center architecture at the current step and thus easily get stuck in a local optimum (See results in Figure~\ref{fig:comparisons_with_rewards}).
To avoid this, we encourage the exploration ability by maximizing the performance improvement between $\beta$ and the center architecture $\alpha$ in $\Omega_{\alpha}$, \ie, $R(\beta|\alpha) {=} R(\beta,w_{\beta}) {-} R(\alpha,w_{\alpha})$,
where $R(\alpha, w_{\alpha})$ denotes some performance metric (\eg, validation accuracy) and $w_{\alpha}$ denotes the optimal parameters of $\alpha$ trained on some dataset.
This can be thought of as finding the subspace with the largest potential to find better architectures, instead of the one containing the best architecture found previously.
Formally, we seek to solve the following optimization problem:
\begin{equation}\label{eq:multiple_anchor_objective}
	\begin{aligned}
	\max_{\pi_{G},\pi_{L}}~\mmE_{\alpha \sim \pi_{G}} \left[\mmE_{\Delta \alpha \sim \pi_{L}}  R(\beta|\alpha) \right], ~~~\text{s.t.}~\beta=\alpha {\oplus} \Delta \alpha
	\end{aligned},
\end{equation}
Since we would update the searched subspace with the resultant architecture $\beta$ (See Figure~\ref{fig:overview}), the performance improvement of the previously searched subspace may not always be the largest one and our method is able to explore other subspaces in the subsequent iterations.

\subsection{Global Search with Dominative Subspace Mining}
\label{sec:global_search}

As the first step of \sexyname,
we seek to mine dominative subspaces via a global search process. Specifically, we first discuss the construction of candidate subspaces. Then, we describe the details of our global search algorithm. 

\begin{algorithm}[t]
	\caption{\small{Training method for \sexyname}.}
    	\begin{algorithmic}[1]\small
            \REQUIRE Search space $\Omega$, global policy $\pi_G(\cdot;\theta_G)$ and local policy $\pi_L(\cdot;\theta_L)$.
            \STATE Train the parameters of the supernet.
            \STATE Randomly sample architectures $\{\alpha_i\}_{i=1}^{K}$ from $\Omega$ to build the subspaces $\{\Omega_{\alpha_i}\}_{i=1}^{K}$ using Eqn.~(\ref{eq:subspace}).
            \STATE Construct the subspace graph $\mG$ based on $\{\Omega_{\alpha_i}\}_{i=1}^{K}$.
            \WHILE{not convergent}
                \STATE \emph{// Perform global search to mine dominative subspace $\Omega_{\alpha}$}
                \STATE Sample a subspace $\Omega_{\alpha} \sim \pi_{G}(\cdot | \mG; \theta_{G})$ centered on $\alpha$.
                \STATE \emph{// Perform local search in the mined subspace $\Omega_{\alpha}$}
                \STATE Sample modifications $\Delta \alpha \sim \pi_{L}(\cdot | \Omega_{\alpha}; \theta_{L})$.
                \STATE Build a resultant architecture $\beta = \alpha \oplus \Delta\alpha$.
                \STATE Compute reward $R(\beta|\alpha)\small{=}R(\beta,w_{\beta}) \small{-} R(\alpha,w_{\alpha})$ using the weights inherited from the supernet. 
                \STATE \emph{// Update subspaces with the locally searched architecture $\beta$}
                \IF{$R(\beta|\alpha) > 0$}
                    \STATE Replace the candidate subspace $\Omega_{\alpha}$ with $\Omega_{\beta}$.
                    \STATE Update the edges connected to $\Omega_{\beta}$ in $\mG$.
                \ENDIF
                \STATE Update the parameters $\theta_G$ and $\theta_L$ by optimizing Eqn.~(\ref{eq:multiple_anchor_objective}) using policy gradient~\cite{williams1992simple}.
            \ENDWHILE
    	\end{algorithmic}
		\label{alg:training}
\end{algorithm}

\textbf{{Subspace Construction}}.
At the beginning of our search method, we seek to construct candidate subspaces centered on a set of architectures for search.
To this end, we randomly collect a set of discrete architectures $\{\alpha_i\}_{i=1}^{K}$ from the search space and build the candidate subspaces $\{\Omega_{\alpha_i}\}_{i=1}^{K}$ around them. 
Let $D({\alpha}, \beta)$ be a function to measure the \textit{Architecture Distance} between two architectures $\alpha$ and $\beta$.
Specifically, we take an architecture $\alpha$ as the center of the subspace $\Omega_{\alpha}$ and constrain all architectures $\beta$ in $\Omega_{\alpha}$ to have distances less than a specific threshold $M$ from the center architecture $\alpha$, \ie, $D({\alpha}, \beta) \leq M$.
Formally, we construct the subspace $\Omega_{\alpha}$ \wrt a center architecture $\alpha$ as:
\begin{equation}\label{eq:subspace}
    \Omega_{\alpha}=\{{\beta} ~|~D({\alpha}, \beta) \leq M, \beta \in \Omega \}.
\end{equation}

Note that architectures in different subspaces may have different operations/topologies and different performances.
To exploit the relationship/information among different subspaces, as shown in Figure~\ref{fig:overview}, we build a subspace graph $\mG=(\mV,\mE)$ to guide the search. 
Here, $\mV$ is a set of nodes and each node denotes a specific subspace $\Omega_{\alpha_i}$. 
$\mE$ is a set of directed edges from a weak subspace (with a poor center architecture) to a better subspace (whose center architecture has higher accuracy).
Note that even when initializing the center architectures $\{\Omega_{\alpha_i}\}_{i=1}^{K}$ randomly, our proposed \sexyname is able to achieve good search performance.
We may further improve \sexyname by taking existing well-designed/searched architectures to construct the subspace (See experimental results in Tables~\ref{tab:nasbench_search} and~\ref{tab:imagenet}).

In the subspace graph $\mG$, since the node/subspace $\Omega_{\alpha}$ is uniquely determined by its centered architecture $\alpha$, we use the embedding of the architecture $\alpha$ to represent the subspace $\Omega_{\alpha}$.
Specifically, for architecture $\alpha$, we use the concatenation of the learnable vector of each component in it to represent the corresponding embedding $\mathbf{h}_{\alpha}$.
For each edge, we represent its embedding $\mathbf{e}_{\alpha\alpha'}$ by $\mathbf{h}_{\alpha'} - \mathbf{h}_{\alpha}$.
The edge in the subspace graph implies how to modify an architecture to obtain another, \eg, replacing convolution with max pooling in some layer.
Given two center architectures, if the components are the same in some positions, the corresponding part in the edge features will be zero.
In this case, the edges explicitly capture the information on how to modify one architecture to another, serving as a rich source of information for the local search and providing exemplary guidance on how to enhance an architecture via modification. To verify the significance of the subspace graph, we conduct ablation studies in Section~\ref{sec:ablation_subspace_graph}. Empirical evidence clearly demonstrates that \sexyname with the subspace graph significantly outperforms its counterpart without this feature.

\textbf{Searching for Dominative Subspaces}.
In each search step, we conduct a global search to automatically mine/select a dominative subspace $\Omega_{\alpha}$ from all candidate subspaces $\{\Omega_{\alpha_i}\}_{i=1}^{K}$. As mentioned before, we seek to find $\Omega_{\alpha}$ that has the potential to achieve a large performance improvement $R(\beta|\alpha)$. 
(Eqn.~(\ref{eq:multiple_anchor_objective})). 
Here, $\alpha$ is the center architecture of $\Omega_{\alpha}$ and $\beta$ is another architecture in this subspace. To this end,  
we devise a controller model that contains a two-layer graph neural network (GNN)~\cite{you2020handling} and an LSTM network.
Specifically, to exploit the information of the subspace graph $\mG=(\mV,\mE)$,
we first employ the GNN to extract features from $\mG$.
Then, we feed the extracted features into the LSTM that samples a candidate subspace $\Omega_\alpha$ via a classifier.

Note that the search performance greatly depends on the subspaces, if we fix all candidate subspaces in $\mG$ during search. In this sense, the controller may get stuck in a local optimum due to the very limited search space
(See results in Figure~\ref{fig:all_comparisons_subspace_updating}).
To address this issue, we propose a simple strategy to gradually update/improve the candidate subspaces $\{\Omega_{\alpha_i}\}_{i=1}^{K}$ using the newly searched architectures.
Specifically, for a selected subspace $\Omega_{\alpha}$ (See Figure~\ref{fig:overview} and lines 12-15 in Algoritm~\ref{alg:training}), we replace its center architecture $\alpha$ with the locally searched architecture $\beta$ if $\beta$ yields better performance than $\alpha$.
This follows the hypothesis that the subspace around a better architecture may be more likely to contain promising architectures.
Once the center architecture is updated, the corresponding subspace $\Omega_{\alpha}$ is also updated to $\Omega_{\beta}$.
Then, we also update the subspace graph $\mG$ by updating all the edges that are originally connected to $\Omega_{\alpha}$.
In this way, the candidate subspaces constantly improve, which helps to explore more and more promising spaces.
After the search, we select the best center architecture in the subspace graph as the inferred architecture according to the validation performance.

\begin{figure}[t]
\centering
\includegraphics[width=\linewidth]{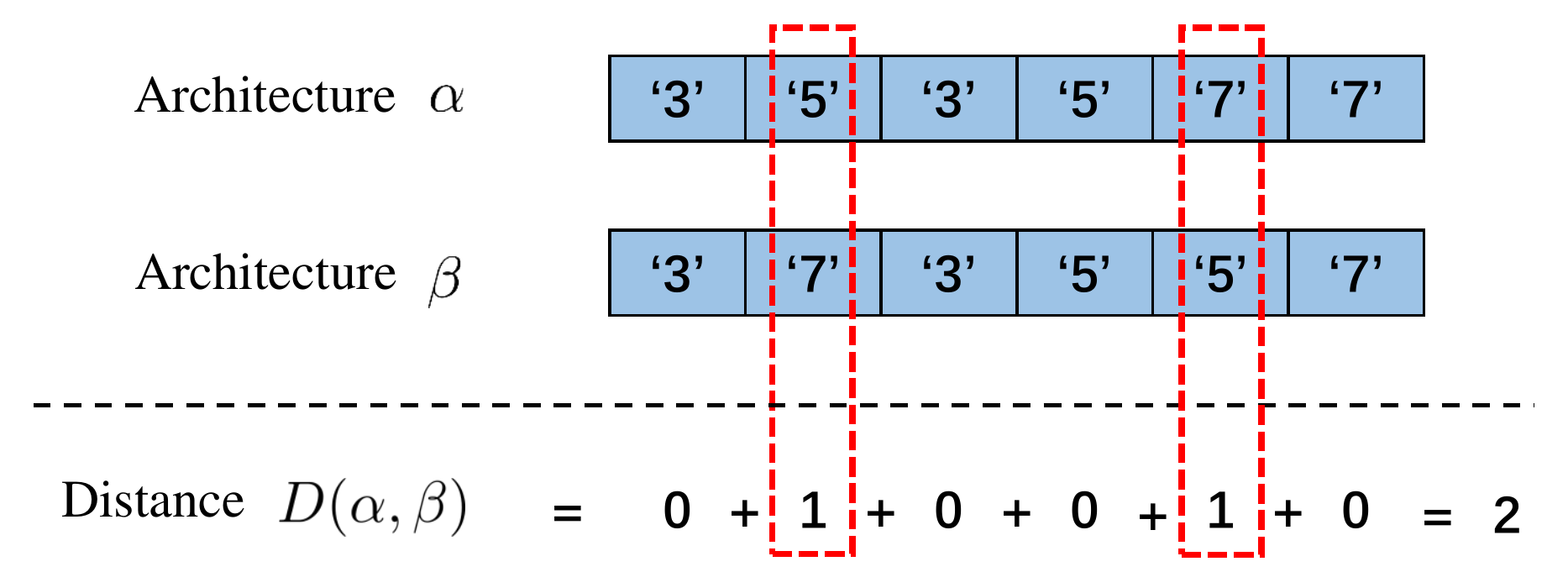}
\caption{An illustration of the architecture representation method and calculation of the architecture distance.
We represent architecture as a string, in which each item denotes an operation (\eg, convolution).
For example, `3', `5' and `7' denote $3\times3$, $5\times5$ and $7\times7$ convolution, respectively.
}
\label{fig:arch_representation}
\end{figure}

\subsection{Local Search in the Mined Subspace}
\label{sec:local_search}

Given a searched subspace $\Omega_{\alpha}$,
we perform local search to find effective architectures.
To guarantee that the search process is limited to the subspace,
we first discuss how to measure the distance between architectures. 
Then, we describe the details of our local search algorithm.

Before defining the \emph{Architecture Distance} between two architectures, we first revisit the representation of architectures, making it easier to understand.
Following~\cite{Cai2020Once,pham2018efficient}, we represent an architecture as a $L$-dimensional string $\alpha \small{=} [\alpha^{(1)}, \alpha^{(2)}, \cdots, \alpha^{(L)}]$, where $L$ is the number of components in the architecture and each $\alpha^{(i)}$ denotes some operation (\eg, convolution).
We propose to compute the \emph{Architecture Distance} $D(\cdot,\cdot)$ by counting the number of different components between two architectures (See Figure~\ref{fig:arch_representation}).
Note that this distance metric is able to calculate the distance between two architecture with different depths.
For non-existent layers in the architecture, we use a placeholder value (like ``0") to ensure that the strings are of the same length.
Let $\mathbbm{1}\{\cdot\}$ be the indicator function.
Given two architectures $\alpha$ and $\beta$, the distance between them is

\begin{align}\label{eq:architecture_distance}
    D(\alpha,\beta) := \sum_{i=1}^{L} \mathbbm{1}\{\alpha^{(i)} \neq \beta^{(i)}\}.
\end{align}

We empirically demonstrate that the proposed distance metric is reasonable and effective when considering all components in the architecture as equal. 
This empirical relationship between accuracy and architecture distance is shown in Figure~\ref{fig:acc_vs_dist}.
We observe that architectures with smaller distances tend to have similar performance, thereby confirming the reliability of the proposed distance metric.
Interestingly, a similar phenomenon has also been observed in well-known NAS methods~\cite{ying2019bench,guo2020breaking}.
This consistency strengthens our justification for the effectiveness of the proposed distance metric.
Consequently, when we identify a dominant subspace centered around a high-performing architecture, it becomes considerably easier to discover better architectures through local search techniques.

To ensure that the locally searched architecture $\beta$ belongs to $\Omega_{\alpha}$,
following common practices~\cite{pham2018efficient,guo2020breaking},
we adopt an LSTM model with a local policy $\pi_L$ that modifies the center architecture $\alpha$ $M$ times.
Specifically, for each time modification, the local policy $\pi_L$ determines which layer has to be modified and which kind of operation is to be applied to this layer. 
Then, by applying $\Delta\alpha$ to $\alpha$, we obtain the modified/searched architecture $\beta=\alpha\oplus\Delta\alpha$ in the subspace. 
Such a decision process is as repeated $M$ times to obtain the complete modification $\Delta\alpha$.
The modification process can be considered as a sequential decision-making process.
This sequential generation process aligns perfectly with the capabilities of LSTM models, which make predictions based on previously determined results.
Note that if $\pi_L$ selects the same operation as the original one in some layers, it will produce no modification to the architecture. Thus, after $M$ times single-layer modification, the resultant architecture $\beta$ still belongs to $\Omega_{\alpha}$, \ie, satisfying the constraint $D(\alpha, \beta) \leq M$.

\textbf{Analysis of the size of local search space.}
Let $L$ be the number of components in an architecture and $C$ be the number of candidate operations for each component.
Traditional NAS methods directly search in the whole search space, whose size is $|\Omega|=C^L$.
In our \sexyname, given a local search distance $M$, the size of the subspace becomes $|\Omega_{\alpha}| {=} \binom{L}{M} C^M$, where $\binom{\cdot}{\cdot}$ denotes the combination function. 
When $M \small{\ll} L$, the subspace would be much smaller than the whole space. In this case, the union of these subspaces constructed with a small $M$ may not cover the entire search space.
Nevertheless, it is exactly our key idea that we seek to focus on some promising subspaces instead of the entire search space to enhance both the search performance and search efficiency.
In practice, taking MobileNet-like search space as an example, we have $C=9$, $L=20$ and $M=3$. The size of the whole search space is $9^{20}$. In contrast, the size of our search subspace is $20!/(3!\times 17!)\times9^3=1140\times9^3$, which is much smaller the the size of the whole space.
In the extreme case, when we consider $M\small{=}L$, the subspace is exactly the whole space and our method reduces to the standard NAS.
We investigate the effect of $M$ on the performance of \sexyname in Section~\ref{sec:ablation_on_M}.

\begin{table*}[t]
	\centering
	\caption{Comparisons with existing methods in NAS-Bench-201~\cite{dong2020nasbench201}.
	``ImageNet-16-120'" denotes a subset of the ImageNet dataset with 120 classes and 16$\times$16 resolution.
	''Search Cost" denotes the time cost in the search phase (measured by second). $^\dagger$ denotes we implement the baselines with the official code under our same settings. Our \sexyname achieves higher accuracy than state-of-the-art methods with less search cost, which verifies the search performance and efficiency of our method.}
    {
    \resizebox{0.75\linewidth}{!}{
	\begin{tabular}{c|c|ccc}
		\topline
		Method & Search Cost (s) & CIFAR-10 & CIFAR-100  & ImageNet-16-120 \\
		\midline
		Random Search$^\dagger$ & 25k & 93.88\scriptsize$\pm$0.27 & 71.54\scriptsize$\pm$1.04 & 45.19\scriptsize$\pm$1.06 \\
		REINFORCE~\cite{williams1992simple}$^\dagger$ & 25k & 93.85\scriptsize$\pm$0.37 & 71.71\scriptsize$\pm$1.09 & 45.24\scriptsize$\pm$1.18 \\
		PNAS~\cite{liu2018progressive}$^\dagger$ & 25k & 93.71\scriptsize$\pm$0.29 & 70.89\scriptsize$\pm$0.99 & 44.75\scriptsize$\pm$0.80\\
		CNAS~\cite{guo2020breaking}$^\dagger$ & 25k & 93.95\scriptsize$\pm$0.28 & 71.73\scriptsize$\pm$1.05 & 45.46\scriptsize$\pm$0.97 \\
		ENAS~\cite{pham2018efficient} & -- & 53.89\scriptsize$\pm$0.58 & 13.96\scriptsize$\pm$2.33 & 14.84\scriptsize$\pm$2.10 \\
		DARTS~\cite{liu2018darts} & 30k & 54.30\scriptsize$\pm$0.00 & 15.61\scriptsize$\pm$0.00 & 16.32\scriptsize$\pm$0.00 \\
		SETN~\cite{dong2019oneshot} & 34k & 87.64\scriptsize$\pm$0.00 & 59.05\scriptsize$\pm$0.24 & 32.52\scriptsize$\pm$0.21 \\
		GDAS~\cite{dong2019searching} & 32k & 93.61\scriptsize$\pm$0.09 & 70.70\scriptsize$\pm$0.30 & 41.71\scriptsize$\pm$0.98 \\
		DSNAS~\cite{hu2020dsnas} & -- & 93.08\scriptsize$\pm$0.13 & 31.01\scriptsize$\pm$16.38 & 41.07\scriptsize$\pm$0.09 \\
		PC-DARTS~\cite{xu2020pcdarts} & -- & 93.41\scriptsize$\pm$0.30 & 67.48\scriptsize$\pm$0.89 & 41.31\scriptsize$\pm$0.22 \\
        DARTS-~\cite{chu2021dartsminus} & 30k & 93.80\scriptsize$\pm$0.07 & 73.02\scriptsize$\pm$0.16 & 46.41\scriptsize$\pm$0.14 \\
        EZNAS~\cite{yash2022eznas} & -- &  93.63\scriptsize$\pm$0.12 & 69.82\scriptsize$\pm$0.16 & 43.47\scriptsize$\pm$0.20 \\
        GradSign~\cite{zhang2022gradsign} & -- & 93.31\scriptsize$\pm$0.47 & 70.33\scriptsize$\pm$1.28 & 42.42\scriptsize$\pm$2.81 \\
        ZiCO~\cite{li2023zico} & -- & 93.50\scriptsize$\pm$0.18 & 70.62\scriptsize$\pm$0.26 & 42.04\scriptsize$\pm$0.82 \\
		\midline
		\sexyname (Ours) & 25k & \textbf{94.23\scriptsize$\pm$0.22} & 72.76\scriptsize$\pm$0.80 & 46.13\scriptsize$\pm$0.67 \\
		\sexynameplus (Ours) & 25k & -- & \textbf{73.12\scriptsize$\pm$0.61} & \textbf{46.66\scriptsize$\pm$0.52} \\
		\bottomline
	\end{tabular}
	}
	}
	\label{tab:nasbench_search}
\end{table*}

\section{Experiments}

The experiments are organized as follows.
We first perform experiments in NAS-Bench-201~\cite{dong2020nasbench201} search space to demonstrate the effectiveness of our \sexyname.
Then, we evaluate our \sexyname in MobileNet-like~\cite{howard2019searching} search space and compare the performance of our method with SOTA methods on a large-scale benchmark dataset ImageNet~\cite{deng2009imagenet}.
Our code and the pretrained models are publicly available at \href{https://github.com/chenyaofo/DSM-NAS}{https://github.com/chenyaofo/DSM-NAS}.

\subsection{Performance Comparisons on NAS-Bench-201}

\textbf{Search Space.}
We apply our \sexyname to a cell-based NAS-Bench-201 search space~\cite{dong2020nasbench201}.
Each cell is a directed acyclic graph with 4 nodes and 6 edges.
Each edge is associated with an operation, which has 5 different candidates, including \textit{zeroize}, \textit{skip connection}, \textit{1$\times$1 convolution}, \textit{3$\times$3 convolution} and \textit{3$\times$3 average pooling}.
Since we search for the candidate operation for each edge, there are $5^6=15625$ candidate architectures in total.
For each architecture, NAS-Bench-201 provides precomputed training, validation, and test accuracies on three different datasets, namely CIFAR-10, CIFAR-100 and ImageNet-16-120. Note that ImageNet-16-120 is a subset of ImageNet~\cite{deng2009imagenet} dataset with 120 classes and 16$\times$16 image resolution.

\textbf{Implementation Details.}
Following the settings in NAS-Bench-201~\cite{dong2020nasbench201}, we use the validation accuracy in epoch 12 as the reward and report the test accuracy in epoch 200 to compare with other baseline methods.
For a fair comparison, we consider the evaluation time of candidate architectures when computing the search cost (limited to 25k seconds).
Following the setting in ~\cite{pham2018efficient}, we train our \sexyname with a batch size of 1 and set the strength of the entropy regularizer to $7.5 \times 10^{-4}$.
We use an Adam optimizer with a learning rate of $1 \times 10^{-2}$.
We set the number of candidate subspaces $K$ to 4 and the search distance $M$ to 4.

To simplify implementation, we combine global policy $\pi_G$ and the local policy $\pi_L$ into a single policy to make decisions to predict the dominative subspace $\Omega_\alpha$ and the modification $\Delta \alpha$.
In other words, we seek to learn a joint policy that first selects a candidate subspace and then finds a promising architecture modification in the selected subspace.
Note that combining the global and local policies together is equivalent to treating them individually.
The publicly available source code demonstrates that the implementation effort required for our DSM-NAS is comparable to that of traditional reinforcement learning-based NAS methods.

\textbf{Comparisons with State-of-the-art Methods.}
We compare our method with two baselines, namely \textit{Random Search} and \textit{REINFORCE}~\cite{williams1992simple}. 
Random Search randomly samples architectures and selects the one with the highest accuracy among them as the final architecture.
REINFORCE performs search by directly maximizing the expectation of the performance of sampling architectures with reinforcement learning.
We report the average test accuracy on three datasets over 500 runs with different seeds.
In this experiment, we initialize the subspace graph with randomly sampled centered architectures.
We also conduct experiments using the subspace graph initialized with searched well-designed architectures, which achieves better search performance.

From Table~\ref{tab:nasbench_search}, compared with the baselines, our \sexyname achieves the highest average accuracy on three datasets, \ie, CIFAR-10, CIFAR-100 and ImageNet-16-120.
Compared with REINFORCE~\cite{williams1992simple} baseline, the proposed \sexyname yields better search accuracy and lower variance (\eg, 72.76$\pm$0.80\% \vs 71.71$\pm$1.09\% on CIFAR-100).
In addition, our \sexyname consistently outperforms the best competitor DARTS- across all three datasets.
This is because our \sexyname searches by focusing on the small but effective subspace, which reduces the search difficulty resulting from the large search space.
During the search, \sexyname updates the candidate subspaces with the locally searched architectures and finds better architectures in the constantly improved subspaces.
Besides, our \sexyname not only delivers superior performance but also achieves them with greater efficiency. It requires only 25k seconds for the search process, compared to DARTS~\cite{liu2018darts} (30k) and DARTS-~\cite{chu2021dartsminus} (30k), and GDAS~\cite{dong2019searching} (32k). This significant reduction in search time underlines the methodological efficiency of \sexyname, which concentrates on mining potent subspaces automatically, rather than combing through the entire search space exhaustively.

\textbf{Transferability of Subspace Graph to New Datasets.}
When we search on a new target dataset, we often have to search from scratch (\ie, initializing candidate subspaces with randomly sampled architectures).
Instead of initiating the search from scratch with randomly sampled architectures, we can leverage promising architectures from existing datasets to initialize candidate subspaces. This variant of our method, called \sexynameplus, not only potentially reduces computational costs but also enhances performance, ensuring a more efficient and effective search process.
To verify this, we conduct experiments on NAS-Bench-201 by considering CIFAR-10 as the existing dataset and CIFAR-100 as well as ImageNet-16-120 as two new target datasets. Since CIFAR-10 is the original/existing dataset, conducting a direct evaluation on CIFAR-10 is meaningless. The "-" symbol underscores CIFAR-10's role as a source dataset used to inform our search strategy, rather than as a target for performance evaluation.

From the results in Table~\ref{tab:nasbench_search}, \sexynameplus with the subspaces transferred from CIFAR-10 achieves higher accuracy and lower variance than \sexyname without that (73.12{$\pm$0.61}\% \vs~72.76{$\pm$0.80}\% on CIFAR-100, 46.66{$\pm$0.52}\% \vs~46.13{$\pm$0.67}\% on ImageNet-16-120).
This is because that the well-designed architectures on CIFAR-10 may also have high performance on the other datasets (\eg, CIFAR-100 and ImageNet-16-120).
Thus, subspaces searched in CIFAR-10 provide a good initialization when searching in a new dataset.
Employing additional data from CIFAR-10 gives \sexyname a strategic advantage. This approach is particularly useful when adapting to a new target dataset.
This strategy not only potentially lowers computational costs but also boosts performance, ensuring a more efficient and productive search.

\subsection{Performance Comparisons on NAS-Bench-360}
We further apply \sexyname to NAS-Bench-360~\cite{tu2022nasbench360} benchmark. In addition to the image classification task, we further consider the tasks of electromyography signals classification and partial differential equations (PDEs) solving. We first provide the details of tasks and implementation and then compare our \sexyname with state-of-the-art methods.

\begin{table}[t]
	\centering
	\caption{
Comparisons with existing methods in NAS-Bench-360~\cite{tu2022nasbench360}.
We report the classification error (\%) and mean square error in the NinaPro and Darcy Flow, respectively. In both tasks, smaller metrics represent better performance.
For a fair comparison, the search cost of all the methods is limited to 25k seconds. Thus we do not report the search cost. }
 
    {
    \resizebox{0.9\linewidth}{!}{
	\begin{tabular}{c|ccc}
		\topline
		Method &  NinaPro $\downarrow$ & Darcy Flow $\downarrow$ \\
		\midline
        Random Search & 8.09\scriptsize$\pm$0.71 & 0.0252\scriptsize$\pm$0.006 \\
        
	REINFORCE~\cite{williams1992simple} & 8.07\scriptsize$\pm$0.73 & 0.0247\scriptsize$\pm$0.006 \\

        Evolution~\cite{real2019regularized} & 8.15\scriptsize$\pm$0.85 & 0.0244\scriptsize$\pm$0.006 \\

        BOHB~\cite{falkner2018bohb} & 8.17\scriptsize$\pm$0.57  & 0.0194\scriptsize$\pm$0.002 \\

        Hyperband~\cite{li2017hyperband} & 8.16\scriptsize$\pm$0.57 &0.0191\scriptsize$\pm$0.002 \\

        ENAS~\cite{pham2018efficient} & 11.56\scriptsize$\pm$1.12 & 0.253\scriptsize$\pm$0.000 \\
        
        DARTS~\cite{liu2018darts} & 22.06\scriptsize$\pm$2.00 & 0.150\scriptsize$\pm$0.093 \\
    
        RSWS~\cite{li2019random} & 9.82\scriptsize$\pm$1.49 & 0.221\scriptsize$\pm$0.045 \\
    
        GDAS~\cite{dong2019searching} & 17.61\scriptsize$\pm$6.39 & 0.180\scriptsize$\pm$0.103 \\
    
        SETN~\cite{dong2019oneshot} & 14.56\scriptsize$\pm$7.30 & 0.253±\scriptsize$\pm$0.000 \\
        EZNAS~\cite{yash2022eznas} &  7.34\scriptsize$\pm$0.32 & 0.0242\scriptsize$\pm$0.022 \\
        GradSign~\cite{zhang2022gradsign} &  7.65\scriptsize$\pm$0.86 & 0.0228\scriptsize$\pm$0.085 \\
        ZiCO~\cite{li2023zico} & 7.49\scriptsize$\pm$0.25 &  0.0223\scriptsize$\pm$0.043 \\

		\midline
		\sexyname (Ours) &  \textbf{6.76\scriptsize$\pm$0.12} & \textbf{0.0211\scriptsize$\pm$0.030} \\
		\bottomline
	\end{tabular}
	}
	}
	\label{tab:nasbench360_search}
\end{table}

\textbf{Considered Tasks}. Following NAS-Bench-360, we consider two more tasks, \ie, \textit{NinaPro} and \textit{Darcy Flow}. NinaPro seeks to classify hand gestures indicated by electromyography signals. We use a subset of the NinaPro DB5 dataset~\cite{cote2019deep} that contains EMG signals from 10 test individuals with 18 different hand gestures. Darcy Flow is a regression task that focuses on mapping from the initial conditions of a PDE to the solution at a later timestep. Its input is a 2d grid specifying the initial conditions of a fluid, and the output is a 2d grid specifying the fluid state at a later time, with the ground truth being the result computed by a traditional solver. We use the same Darcy Flow dataset that was used in~\cite{li2020fourier}. We report the mean square error. Note that NAS-Bench-360 provides precomputed accuracies on these datasets.

\textbf{Implementation Details}. Following the settings in NAS-Bench-360, the search space is also the same as NAS-Bench-201. The hyperparameters for policy learning are the same as those in NAS-Bench-201. Specifically, we train our \sexyname with a batch size of 1 and set the strength of the entropy regularizer to $7.5 \times 10^{-4}$.
We use the Adam optimizer with a learning rate of $1 \times 10^{-2}$.
We set the number of candidate subspaces $K$ to 4 and the search distance $M$ to 4.

\textbf{Comparisons with State-of-the-art Methods}.
In Table~\ref{tab:nasbench360_search}, we compare our \sexyname on the \textit{NinaPro} and \textit{Darcy Flow} tasks of the NAS-Bench-360 benchmark. We report the average test accuracy over 500 runs with different seeds. Our \sexyname achieves high accuracy than REINFORCE~\cite{williams1992simple} (0.0211 \vs~0.0252 on Darcy Flow) and Evolution~\cite{real2019regularized} (0.0211 \vs~0.0247 on Darcy Flow) baselines.
Compared with the weight-sharing based NAS methods (\eg, ENAS~\cite{pham2018efficient} and DARTS~\cite{liu2018darts}), \sexyname still outperforms them.
This superior performance can be attributed to \sexyname's focus on a smaller, more effective subspace of the search space, as opposed to exploring the entire vast search space. This approach reduces the complexity of the search, making it more likely to yield promising architectural designs.
The results verify the effectiveness of \sexyname across various tasks, extending beyond the realm of image classification.

In addition, we find that weight-sharing NAS methods (\ie, ENAS~\cite{pham2018efficient}, DARTS~\cite{liu2018darts}, RSWS~\cite{li2019random}, GDAS~\cite{dong2019searching} and SETN~\cite{dong2019oneshot}) achieve worse performance than the Random Search baseline on the NinaPro dataset. The reasons may be twofold.
First, the primary issue with weight sharing in methods like ENAS~\cite{pham2018efficient} is multi-model forgetting~\cite{benyahia2019overcoming,zhang2020overcoming}, where the performance of previously trained models deteriorates due to the overwriting of shared parameters during sequential training of multiple networks. The degree of performance degradation is directly proportional to the amount of shared weights, with more sharing leading to greater impacts. Additionally, the complexity of optimization increases, making it more challenging to improve subsequent models without negatively affecting the performance of earlier ones. These challenges underscore the need for careful parameter management to minimize performance interference between models. Therefore, weight sharing cannot provide accurate model performance evaluation, leading to limited search performance. 
Second, prior studies~\cite{tu2022nasbench360, chu2021dartsminus} demonstrate that the search process in methods like DARTS~\cite{liu2018darts}, unintentionally prefers architectures with a higher number of skip connections. While a balanced number of skip connections can boost performance, prompting NAS algorithms to favor the skip connection operation, an excessive focus on this aspect often leads to less optimal search results. Excessive reliance on skip connections would affect search performance, producing architectures overwhelmed by these operations and reducing their overall effectiveness. This narrows the diversity of the explored architecture space and overfits to a particular pattern, which may not necessarily result in superior performance.

\begin{table*}[t]
	\centering
	\caption{Comparisons of the architectures searched/designed by different methods on ImageNet.
	``--'' means unavailable results.
	``\#Queries``, a widely used metric~\cite{luo2020seminas,yan2020does}, denotes the number of architecture-accuracy pairs queried from supernet or performance predictor during the search. A smaller ``\#Queries`` means the search algorithm is more efficient. Our \sexyname outperforms most human-designed and automatically searched architectures with less search cost and fewer search queries.}
	{
	\resizebox{\textwidth}{!}
	{
    \begin{tabular}{c|ccccccc}
    \topline
    \multicolumn{1}{c|}{\multirow{2}[0]{*}{Search Space}} &
    \multicolumn{1}{c}{\multirow{2}[0]{*}{Architecture}} &
    \multicolumn{2}{c}{Test Accuracy (\%)} &
    \multicolumn{1}{c}{\multirow{2}[0]{*}{\#MAdds (M)}} &
    \multicolumn{1}{c}{\multirow{2}[0]{*}{\#Queries (k)}} &
    \multicolumn{1}{c}{Search Time} \\
    \cline{3-4} &  & \multicolumn{1}{c}{Top-1} & \multicolumn{1}{c}{Top-5} &  & & (GPU days) \\
    \midline
    \multirow{3}*{--} & ResNet-18~\cite{resnet} & 69.8 & 89.1 & 1,814 &  -- & -- \\
     & MobileNetV2 ($1.4\times$)~\cite{sandler2018mobilenetv2} & 74.7 & --  & 585 & -- & --  \\
     & ShuffleNetV2 ($2\times$)~\cite{ma2018shufflenet} & 73.7 & -- & 524 & -- & --  \\
    \midline
    \multirow{2}*{NASNet} & NASNet-A~\cite{zoph2018learning} & 74.0 & 91.6 & 564 & 20  & 1800\\
     & AmoebaNet-A~\cite{real2019regularized} & 74.5 & 92.0 & 555 & 20  & 3150\\
    \midline
    \multirow{7}*{DARTS} & DARTS~\cite{liu2018darts} & 73.1 & 91.0 & 595 &19.5  & 4\\
     & P-DARTS~\cite{chen2019progressive} & 75.6 & 92.6 & 577 &11.7  & 0.3\\
     & CNAS~\cite{guo2020breaking} & 75.4 & 92.6 & 576 & 100 & 0.3\\
     & AlphaX~\cite{wang2019alphax} & 75.5 & 92.2 & 579 & -- & 12\\
     & DARTS-~\cite{chu2021dartsminus} & 74.6 & 92.1 & 547 & 93.6 & 4.5 \\
     & Shapley-NAS~\cite{Xiao_2022_CVPR} & 76.1 & -- & 582 & -- & 4.2 \\
     & RF-DARTS~\cite{Zhang_2023_CVPR} & 76.0 & 92.4 & 593 & 93.6 & -- \\
    \midline
     \multirow{15}*{MobileNet-like} & RLNAS~\cite{zhang2021neural} & 75.6 & 92.6 & 473 & -- & -- \\
     & AtomNAS~\cite{Mei2020AtomNAS} & 75.9 & 92.0  & 367 & 78  & --\\
     & Few-shot NAS~\cite{zhao2021few} & 75.9 &--  & 521 & -- & 11.6 \\
    & FairNAS~\cite{chu2021fairnas} & 77.5 & 93.7 & 392 & 11.2 & 12 \\
    & DNA~\cite{li2020block}& 78.4 & 94.0 & 611 & -- & 32.6 \\
    & FBNetV2~\cite{FBNETV2} & 77.2 & --  & 325 & 11.5 & 25\\
    & EfficientNet-B1~\cite{EfficientNet}& 79.2 & 94.5 & 734 & -- & --\\
    & OFA-Large~\cite{Cai2020Once} & 80.0 & 94.9  & 595 & 20 & 51.7\\
    & Cream-L~\cite{peng2020cream} & 80.0 & 94.7 & 604 & -- & 12\\
    & NEAS-L~\cite{chen2021oneshot} & 80.0 & 94.8 & 574 & -- & $<$13 \\
    & GM~\cite{hu2022generalizing} & 76.6 & 93.0 & 530 & -- & 24.9 \\
    & MAGIC-AT~\cite{xu2022analyzing} & 76.8 & 93.4 & 598 & -- & -- \\
    & NAS-LID~\cite{he2023naslid} & 77.1 & 93.7 & 678 & -- & 2.3 \\
    & ZiCo~\cite{li2023zico} & 79.4 & -- & 603 & & 0.4 \\
    & \sexyname (Ours) & 79.9 & 94.8 & 597 & \textbf{10} & \textbf{0.8}\\
    & \sexynameplus (Ours) & \textbf{80.2} & \textbf{94.9} & 582 & \textbf{10} & 51.7+0.8\\
    \bottomline
    \end{tabular}
    }
    }
\label{tab:imagenet}
\end{table*}

\subsection{Performance Comparisons on ImageNet}\label{sec:exp_imagenet}

\textbf{Search Space}.
In this experiment, we further evaluate our method on another benchmark search space, \ie, MobileNet-like search space~\cite{howard2019searching}.
The candidate architecture consists of 5 different units and each of them has consecutive layers.
We search for MBConv in each layer with kernel sizes $R$ selected from $\{3, 5, 7\}$, expansion ratios $E$ selected from $\{3, 4, 6\}$ and the number of layers in each unit selected from $\{2, 3, 4\}$.
To encode the architecture, we use a string of length 20, in which the element in the string represents the combination setting of the expansion ratio and the kernel size.
For instance, the element ``0" denotes non-existent layer, ``1" denotes the layer with $E=3$ and $R=3$, ``2" denotes the layer with $E=3$ and $R=4$, and so on.
In this case, we ensure that our proposed method can effectively calculate the distance between architectures with varying depths.

Following~\cite{wu2019fbnet}, we randomly choose 10\% classes from the original dataset as the training set to train the supernet.
We measure the validation accuracy of sub-networks on 1000 validation images sampled from the training set.
We train the supernet with a progressive shrinking strategy~\cite{Cai2020Once} for 90 epochs.
To compute the performance improvement $R(\beta|\alpha)$ with the validation accuracy, we train a predictor to predict the validation accuracy following~\cite{Cai2020Once}. 

\textbf{Implementation Details.}
Following~\cite{pham2018efficient}, we train \sexyname for 10k iterations with a batch size of 1.
We use an Adam optimizer with a learning rate of $3\times10^{-4}$.
To encourage exploration of \sexyname, we add an entropy regularizer to the reward weighted by $1\times10^{-3}$.
We set the number of candidate subspaces $K$ to 10 and the local search distance $M$ to 3.
We report the search cost based on the NVIDIA Tesla V100 GPU.
Following the \textit{mobile setting}~\cite{liu2018darts}, we constraint the number of multiply-adds (\#MAdds) of the searched architecture to be less than 600M.
To achieve this, we update the candidate subspace (line 12 in Algorithm~\ref{alg:training}) with the architecture that has \#MAdds less than 600M.
To accelerate model evaluation, following~\cite{Cai2020Once,lu2020neural}, we first obtain the parameters from the full network of OFA and then finetune them for 75 epochs on the ImageNet training set (containing 128k images). 
We use a batch size of 512 and an SGD optimizer with a learning rate of 0.012.
The learning rate decay follows the cosine annealing strategy with a minimum of 0.001.
We perform data augmentations including horizontally flipping, random crops, color jittering, and AutoAugment.

\begin{figure*}[t]
\centering
\includegraphics[width=0.7\linewidth]{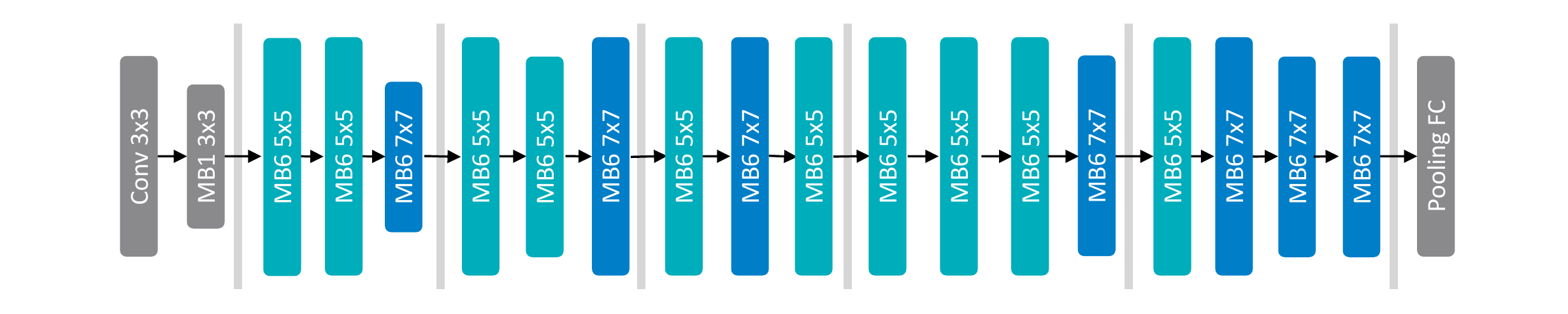}
\caption{
    The architecture searched by \sexyname in MobileNet-like search space.
}
\label{fig:arch_visualization}
\end{figure*}

\begin{figure*}[t]
\centering
\includegraphics[width=0.7\linewidth]{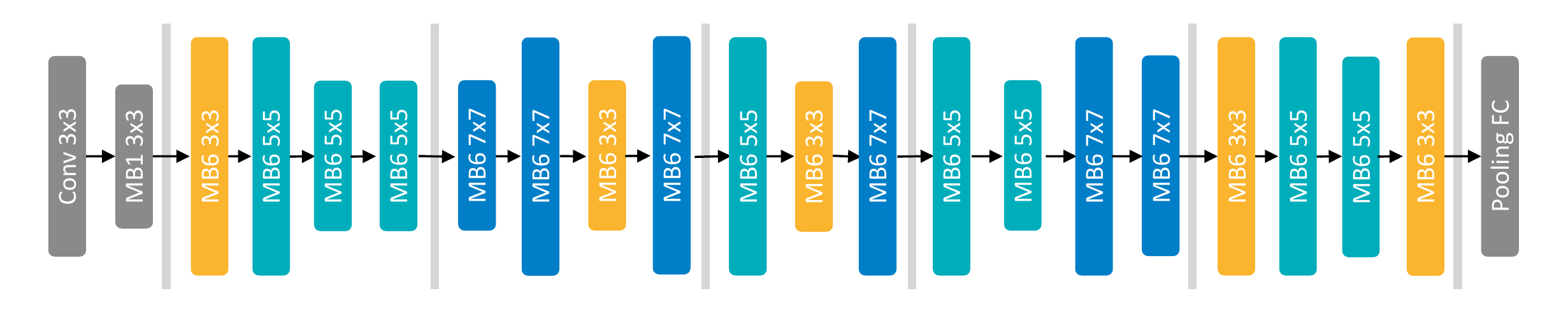}
\caption{
    The architecture searched by \sexynameplus in MobileNet-like search space.
}
\label{fig:arch_visualization_plus}
\end{figure*}

\textbf{Comparisons with State-of-the-art Methods.}
To investigate the effectiveness of the proposed method, we apply our method to MobileNet-like search space as two variants:
1) \textbf{\sexyname} searches based on the subspace graph initialized with randomly sampled architectures, which is suitable in the scenario without any available well-designed architectures.
2) \textbf{\sexynameplus} adopts the subspace graph which is initialized with a set of existing well-designed architectures searched by OFA~\cite{Cai2020Once}.
Note that our \sexyname outperforms SOTA methods and \sexynameplus further boosts the search performance.

As shown in Table~\ref{tab:imagenet}, under the mobile setting, the architecture searched by \sexyname reaches 79.9\% top-1 accuracy and 94.8\% top-5 accuracy, which outperforms not only the manually designed architectures but also most automatically searched ones.
Specifically, \sexyname outperforms the best manually designed architecture (\ie, MobileNetV2) by 5.2\% (\ie, 79.9\% \vs~74.7\%).
Compared with the state-of-the-art NAS method (\eg, OFA and Cream-L), \sexyname also achieves competitive performance (\ie, 79.9\% \vs~80.0\%) with fewer number of queries and search cost (only 0.8 GPU days).
Here, the lower costs mainly benefit from two aspects: 1) accelerating the search policy learning by reducing the number of queries (ours 10k \vs~OFA 20k); 2) accelerating the supernet training by using a proxy dataset (ImageNet-100) and early stop (training 90 epochs).
These results show the effectiveness and efficiency of the proposed \sexyname.

Compared with zero-shot NAS method ZiCo~\cite{li2023zico}, \sexyname achieves higher accuracy than ZiCO (79.9\% \vs~79.4\%) under the \textit{mobile setting} ($\#$MAdds $<$ 600M).
The potential reasons are that our \sexyname searches by focusing on the small but effective subspace, which reduces the search difficulty resulting from the large search space, while ZiCo adopts a vanilla evolutionary algorithm that still searches in the whole search space.
As for the search cost, \sexyname consumes more GPU resources than ZiCO (0.8 \vs~0.4 GPU days).
The reason is that \sexyname needs to train a supernet (0.7 GPU days) on a proxy ImageNet subset, which is costly compared with the zero-shot proxy used in ZiCO.

Compared with \sexyname, our \sexynameplus further improves the top-1 accuracy on ImageNet from 79.9\% to 80.2\%.
Note that \sexynameplus outperforms all of the considered manually-designed and automatically searched architectures.
The reason is that \sexyname searches for the dominative subspace from randomly initialized candidates and needs to improve the candidate subspaces gradually.
Instead, \sexynameplus directly employs the dominative subspaces centered on a set of well-designed architectures at the beginning of the search.
In this case, the architectures in such initialized subspaces have better performance than that in randomly initialized subspaces, which accelerates the search process by providing good subspace initialization centered on the good architectures.
The results demonstrate that we are able to apply our method to existing designed architectures/subspaces to further enhance the search performance.

\textbf{Visualization of Searched Architectures}.
We show the visualization results of \sexyname (top-1 accuracy 79.9\%) and \sexynameplus (top-1 accuracy 80.2\%) in MobileNet-like search space in Figure~\ref{fig:arch_visualization} and Figure~\ref{fig:arch_visualization_plus}, respectively.
From the visualization results, we find that
the number of conv $5\times5$ and conv $7\times7$ is much larger than the number of conv $3\times3$.
The reason may have two aspects. 
First, conv $5\times5$ and conv $7\times7$ have a larger reception field and larger model capacity with more parameters than conv $3\times3$~\cite{tan2019mixconv}. Second, larger conv operations may help the network preserve more information when doing downsampling (mentioned by ProxylessNAS~\cite{cai2018proxylessnas}).
In this case, architectures with conv $5\times5$ and conv $7\times7$ may often be beneficial to achieve better performance than that with conv $3\times3$.
Similar results of "large convs are dominated" can also be found in \cite{Cai2020Once,cai2018proxylessnas}.

\begin{figure*}[t]
\centering
\subfloat[Comparisons in NAS-Bench-201 search space.]{\includegraphics[width=0.9\columnwidth]{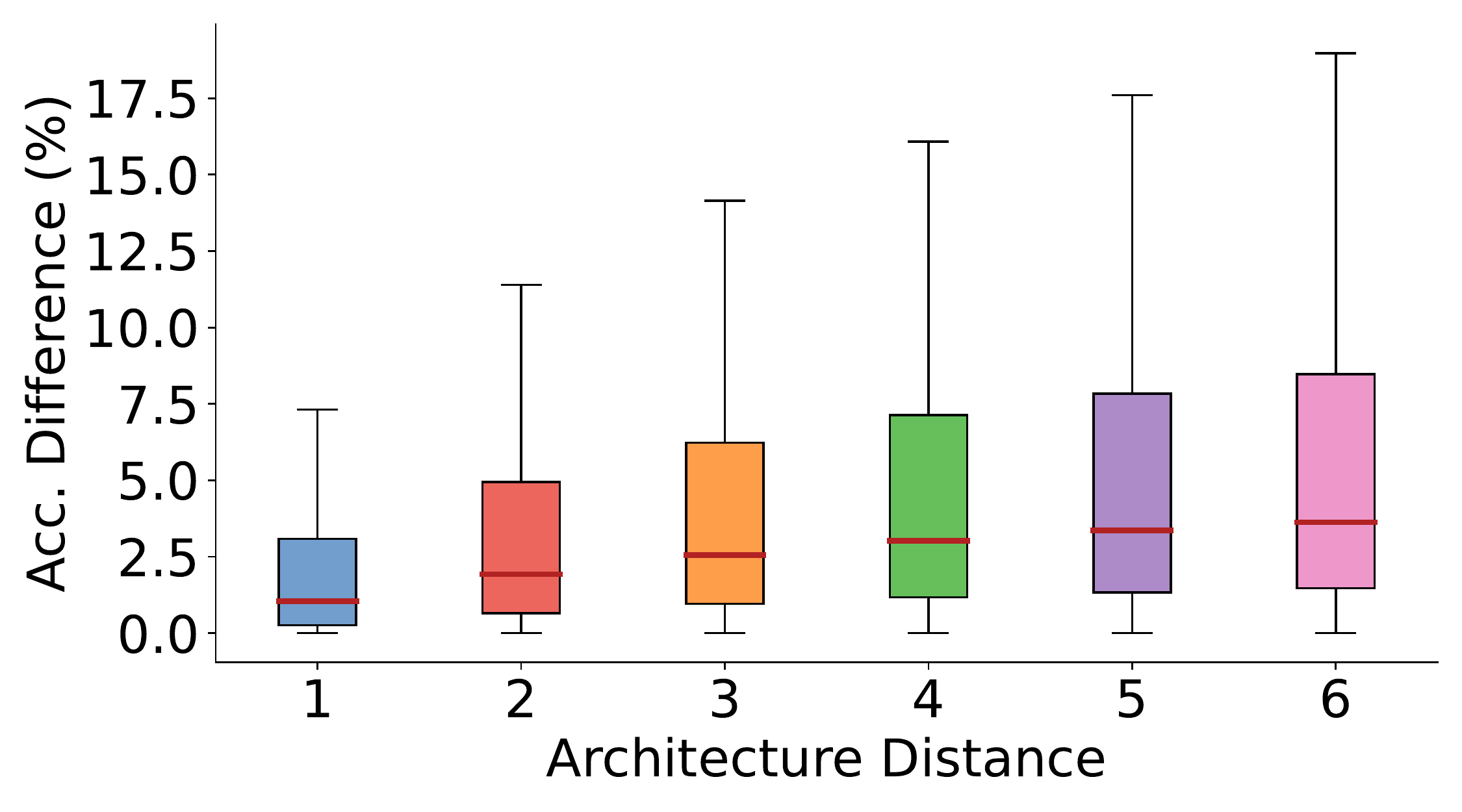}%
\label{fig:acc_vs_dist_nasbench}}
\hfil
\subfloat[Comparisons in MobileNet-like search space.]{\includegraphics[width=0.9\columnwidth]{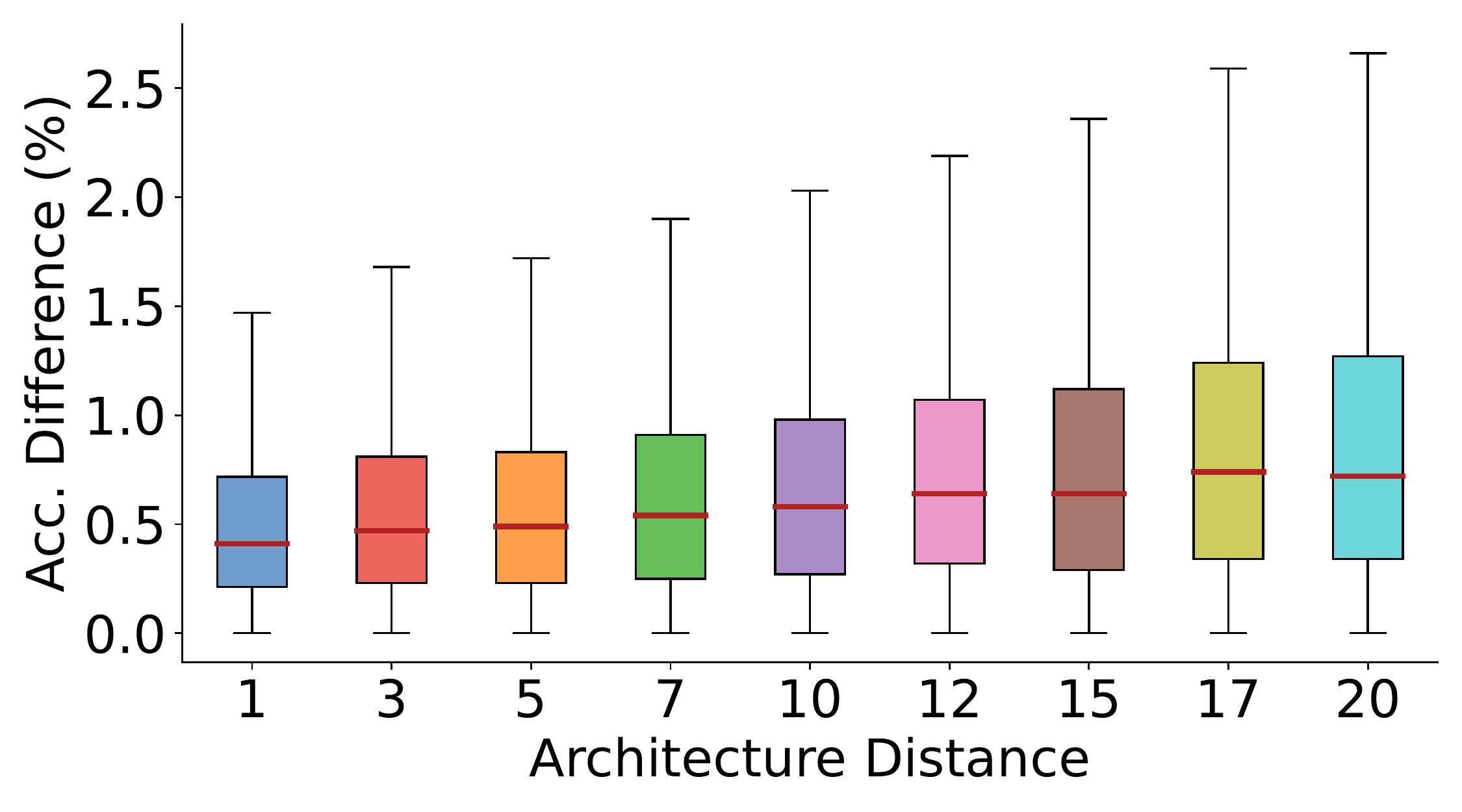}%
\label{fig:acc_vs_dist_mobilenet}}
\caption{The performance difference that is measured by accuracy (\%) \vs~the architecture distance in NAS-Bench-201 search space (a) and MobileNet-like search space (b).}
\label{fig:acc_vs_dist}
\end{figure*}

\section{More Ablation Studies}

In this section, we perform more experiments to demonstrate the effectiveness of the proposed architecture distance (Section~\ref{sec:ablation_arch_distance}), the proposed subspace updating scheme (Section~\ref{sec:ablation_subspace_updating}), the proposed performance improvement reward (Section~\ref{sec:ablation_perf_imp_reward}), and the subspace graph (Section~\ref{sec:ablation_subspace_graph}).
In addition, we conduct ablations to investigate the effect of the number of candidate subspaces $K$ and the local search spaces in Sections~\ref{sec:ablation_on_K} and~\ref{sec:ablation_on_M}, respectively.

\subsection{More Discussions on Architecture Distance}\label{sec:ablation_arch_distance}

As mentioned before, our method is built upon an underlying hypothesis that the neighborhood around a good architecture is usually a dominative subspace.
In other words, the architectures in the neighborhood/subspace are more likely to have good performance.
In this sense, once we find a dominative subspace centered on a good architecture, it would be much easier to find better architectures via a local search.
To build such subspace around a given architecture, we devise an \textit{Architecture Distance} $D(\cdot, \cdot)$ to measure the distance between two architectures.
In the following, we provide empirical results in  NAS-Bench-201 and MobileNet-like search spaces to demonstrate that the devised distance function is able to support the hypothesis.
We conduct experiments in NAS-Bench-201 and MobileNet-like search space by computing the performance difference (measured in accuracy) between two architectures with different distances.

We show the results over 100 different trials in Figure~\ref{fig:acc_vs_dist}.
From the results, the average accuracy difference between two architectures becomes larger when their distance increases.
For example, in NAS-Bench-201 space, the accuracy difference increases from 4.49 to 6.69 when the distance grows from 1 to 2.
Meanwhile, the variance of the performance difference also increases as the corresponding distance grows (\eg, increasing from 6.1 to 9.5 when the distance grows from 1 to 2).
The results demonstrate the rationality and effectiveness of the designed architecture distance, \ie, architectures with smaller distances (in the same subspace) tend to have similar performance.
Thus, we are able to find promising architectures in a dominative subspace around existing good architectures more easily than the overall search space.
Similar observations are also found in NAS-Bench-101~\cite{ying2019bench} search space.

We find that the performance difference in the NAS-Bench-201 search space is more obvious than the MobileNet-like search space. 
The is because that the candidate operations are different in these two search spaces
In the NAS-Bench-201 search space, the available operations are: (1) zeroize, (2) skip connection, (3) 1-by-1 convolution, (4) 3-by-3 convolution, and (5) 3-by-3 average pooling. The "zeroize" operation essentially drops the features, leading to a substantial impact on model performance. When an architecture undergoes a transition that involves the "zeroize" operation, the model performance changes sharply.
Instead, in the MobileNet-like search space, we search for the expansion ratio and kernel size from $\{3, 4, 6\}$ and $\{3, 5, 7\}$, respectively. Unlike the NAS-Bench-201 space, the MobileNet-like search space does not include the "zeroize" operation. As a result, changes in architecture distance tend to produce more subtle variations.

\begin{figure*}[t]
\centering
\includegraphics[width=0.9\linewidth]{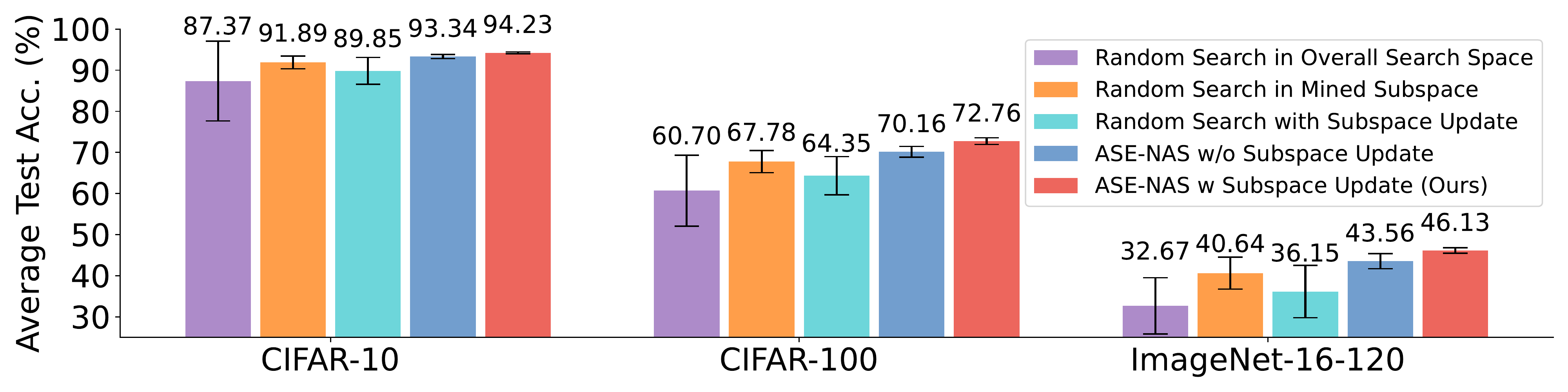}
\caption{
Comparisons of the search performance with/without subspace updating on NAS-Bench-201.
}
\label{fig:all_comparisons_subspace_updating}
\end{figure*}

\subsection{Effect of Subspace Updating Scheme}
\label{sec:ablation_subspace_updating}

In the global search, if we keep all candidate subspaces fixed, the controller may get stuck in a local optimum due to the very limited search spaces, which would lead to poor search performance.
To address this issue, we propose a simple strategy to gradually update the candidate subspaces with the locally searched architectures.
To be specific, we replace the center architecture $\alpha$ in the mined subspace with locally searched architecture $\beta$ if $\beta$ has better performance than $\alpha$.
This simple updating strategy ensures the candidate subspaces would become more and more promising, which helps to find good architecture within the gradually improved subspaces during the local search process.

To verify the effectiveness of the proposed subspace updating scheme, we conduct more experiments on NAS-Bench-201 and compare our \sexyname with three baselines and variants, namely \textit{Random Search in Overall Search Space}, \textit{Random Search in Mined Search Space}, \textit{Random Search with Subspace Update} and \textit{\sexyname without Subspace Update}.
The first two baselines conduct a random search in the entire search space and the mined subspace in the final search step of our method, respectively.
The third baseline performs a random search with our proposed subspace update strategy.
The variant \textit{\sexyname without Subspace Update} uses the same settings as our method but performs a search without updating the candidate subspaces.

From the results in Figure~\ref{fig:all_comparisons_subspace_updating}, random search in the mined subspace (orange) has higher average accuracy and lower variances than that in the overall search space (purple), \ie, 91.89$\pm$1.53\% \vs~87.37$\pm$9.71\%, which demonstrate the effectiveness of the proposed subspace updating scheme.
Our \sexyname with subspace (red) updating outperforms \sexyname without that (blue) on three considered datasets. The results indicate that \sexyname constantly can find dominative subspaces during the search, which also shows the effectiveness of the subspace updating scheme.
In addition, compared with the baseline that searches randomly in the mined subspace, \sexyname consistently achieves better search accuracy.
The experimental results demonstrate that the local search scheme is able to further enhance the search performance by finding promising architectures in the mined subspace.
In addition, the random search with a subspace update baseline (cyan) surpasses the performance of the standard random search across the entire search space (purple). This improvement suggests that updating the search subspace leads to the identification of more promising regions within the search space. However, random search with subspace update (cyan) does not perform as well as random search in a mined search space (orange). This implies that relying solely on random search is insufficient for effectively finding the most effective subspaces.

As for the search speed, random search baselines require less than 1 microsecond to sample architectures from a uniform distribution since it does not depend on any heavy deep models. We sample a total of 100 architectures and then select the one with the highest predicted accuracy. The total time for this process is less than 1 second. For \sexyname, the controller requires 0.1 seconds to generate an architecture by performing global and local searches and 0.22 seconds to update the policy parameters according to the estimated performance by the performance predictor. In MobileNet-like search space, we train the controller for 10k iterations with a batch size of 1 following~\cite{pham2018efficient}. Thus, the policy search cost is 1 GPU hour (\ie, 0.04 GPU days).
Though random search requires a very low search cost, it struggles to uncover promising architectures because of its unplanned attempts. In contrast, \sexyname significantly outperforms random search baselines. This is attributed to our innovative global and local search strategy, which iteratively identifies promising architectures by leveraging knowledge from previous explorations. This strategic method ensures a more directed and efficient search process, leading to superior performance in finding optimal architectures.

\begin{figure*}[t]
\centering
\subfloat[Comparisons with different reward functions.]{\includegraphics[width=0.9\columnwidth]{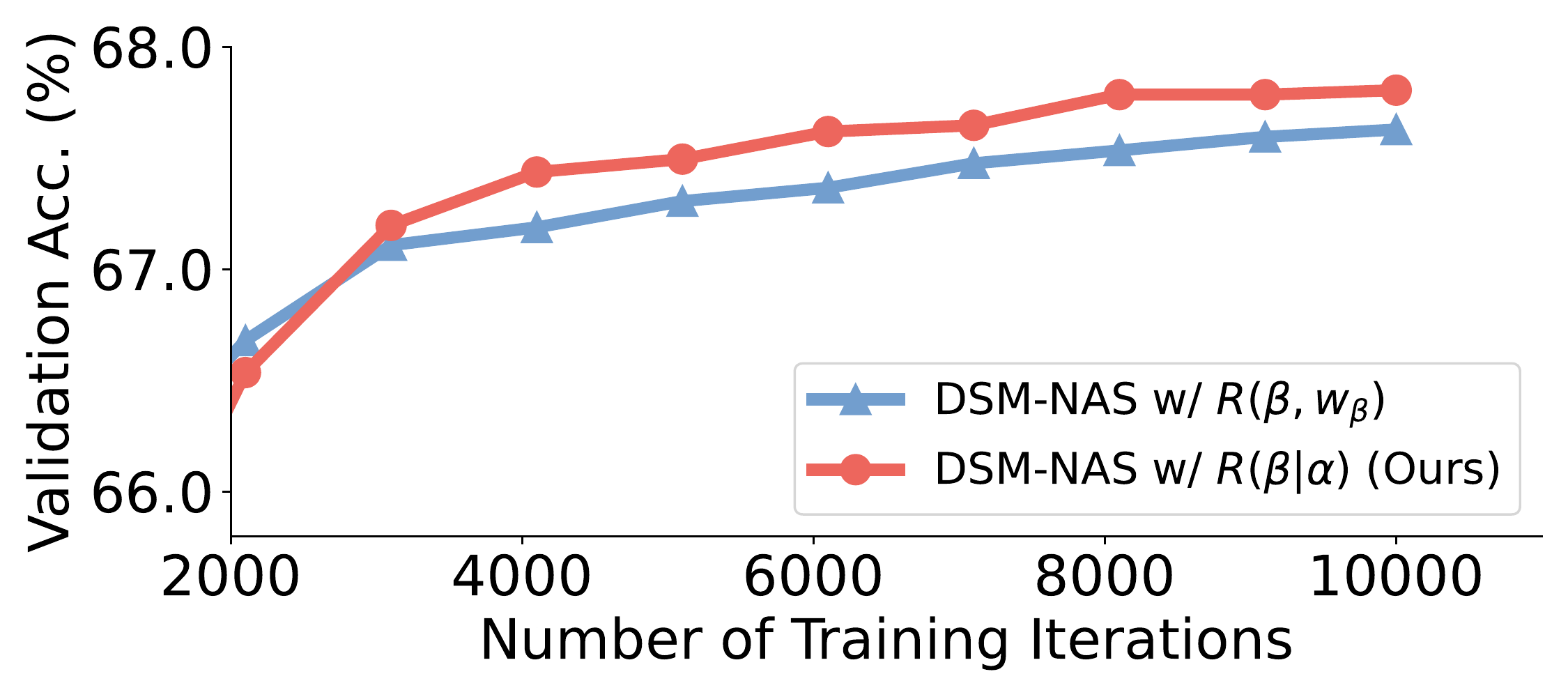}%
\label{fig:comparisons_with_rewards}}
\hfil
\subfloat[Comparisons with/without subspace graph.]{\includegraphics[width=0.9\columnwidth]{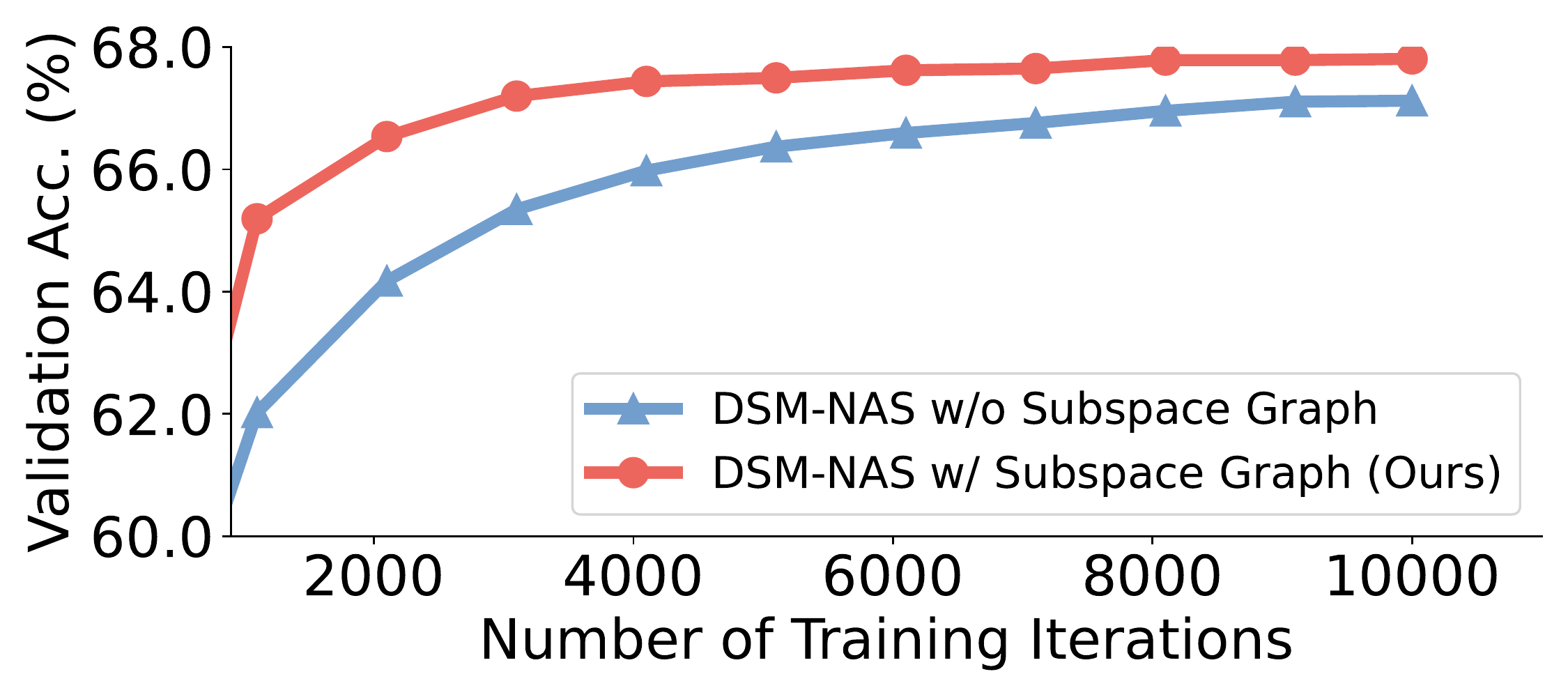}%
\label{fig:comparisons_with_subgraph}}
\caption{Comparisons of the search performance with different reward functions(a) and with/without subspace graph(b) in MobileNet-like search space on ImageNet.}
\label{fig:comparisons_with_rewards_subgraph}
\end{figure*}

\subsection{Effect of the Performance Improvement Reward}\label{sec:ablation_perf_imp_reward}

In policy learning, we use the performance improvement between the resultant $\beta$ and the center architecture $\alpha$ in $\Omega_{\alpha}$ as the reward.
Besides this, one can learn the policy by considering the performance of $\beta$ as the reward.
However, the policy may easily get stuck in a local optimum and always select the same subspace with the highest-performing center architecture.
To verify this, we conduct more experiments in the MobileNet-like search space on ImageNet to compare the search performance with these two kinds of reward functions.
All the hyperparameters are the same as in Section~\ref{sec:exp_imagenet}. We report the validation accuracy over 10 different runs estimated by the accuracy predictor and compare our \sexyname with the variant.
For the tested variant, we maintain consistency in all experimental conditions except for the component under examination.

In Figure~\ref{fig:comparisons_with_rewards_subgraph}(a), 
we employ a variant as the baseline to verify the effectiveness of the proposed reward function.
This variant directly uses the performance of the resultant architecture $\beta$ as reward in policy learning.
From the results, \sexyname using the performance improvement (red) achieves higher search performance than that using the performance of resultant architecture (blue) (67.81$\pm$0.18\% \vs~67.63$\pm$0.31\%).
In addition, maximizing the performance improvement finds more new subspaces than using the absolute performance during the search process (319 \vs~284).
The reason is that if we directly maximize the performance of the resultant architecture $\beta$, the search algorithm may always select the same subspace with the best architecture.
In this case, the remaining subspaces are ignored, which results in poor explorations during the search.
In contrast, using performance improvement encourages explorations among different subspaces.

\subsection{Effect of Subspace Graph}\label{sec:ablation_subspace_graph}

We build the subspace graph with a set of candidate subspaces.
In the graph, nodes denote candidate subspaces and direct edges denote the relationships among them.
In practice, we represent direct edges as a modification vector that implies how to modify the center architecture in the weak subspace to that in the better subspace.
These edges take helpful information for the local search since they are good examples to represent how to modify an architecture to a better one for the local policy.
In addition, the architectures in different subspaces may have different computational operations/topologies and different performances.
In this case, the graph structure in the subspace graph may convey beneficial information to select a dominative subspace in the global search.

We investigate the effect subspace graph by performing more experiments in the MobileNet-like search space with/without the subspace graph structure.
The experimental setup is the same as that in Section~\ref{sec:ablation_perf_imp_reward}.
Specifically, we perform an ablation study by comparing our \sexyname with a variant without the subspace graph structure.
For this variant, we treat the candidate subspaces as separate points and extract the features from them using two fully-connected layers instead of the two-layer graph neural network.
We show the results in Figure~\ref{fig:comparisons_with_rewards_subgraph}(b).
We report the averaged validation accuracy (obtained by the supernet) over 10 different runs.
From the results, \sexyname with subspace graph (red) has not only higher validation accuracy but also lower variance than \sexyname without that (blue) (67.81$\pm$0.18\% \vs~67.12$\pm$0.46\%).
The results demonstrate the significant role of the subspace graph in guiding the search process. Our DSM-NAS with the subspace graph outperformed the variant without it, highlighting the graph's effectiveness in narrowing the search scope and concentrating efforts on promising architectural regions.

\subsection{Effect of the Number of Candidate Subspaces}\label{sec:ablation_on_K}

We build the subspace graph with $K$ architectures and thus have $K$ candidate subspaces $\{\Omega_{\alpha_i}\}_{i=1}^K$.
When we consider a small $K$, the information carried by the subspaces would be limited, resulting in poor search performance.
In contrast, a larger $K$ means more exploration in the candidate subspaces.
Nevertheless, too many candidate subspaces would introduce a heavy computational burden since the computational cost of the GNN increases quadratically as $K$ becomes larger.
To investigate the effect of $K$, we conduct an ablation study with different $K$ on ImageNet.
For fair comparisons, we set the number of each subspace updates in the graph to be the same.

In Figure~\ref{fig:comparisons_hyperparameter}(a), \sexyname achieves the worst validation accuracy when $K\small{=}1$ since it only explores a single subspace during the search process, which greatly depends on the initialized architecture.
As $K$ becomes larger, \sexyname achieves better validation accuracy.
The reason is that more architectures benefit the search by exploring more diverse dominative subspaces.
In this case, \sexyname has a larger probability of finding promising architectures.
Besides, when $K$ is larger than 10, \sexyname yields very similar search performance.
These results demonstrate that using 10 different subspaces is sufficient to achieve competitive performance.
Thus, we set $K$ to 10 on ImageNet.

\subsection{Effect of the Local Search Distance}\label{sec:ablation_on_M}

When performing the local search in $\Omega_\alpha$, we use a hyper-parameter $M$ to restrict the size of the local search subspace.
A smaller search distance $M$ means that we perform the local search in a smaller subspace.
Note that searching in small but effective subspaces is exactly our core idea to enhance both search performance and efficiency.
In contrast, a larger $M$ enables \sexyname to explore more architectures in the search space but makes it harder to explore the whole search space.
Note that the maximum of $M$ equals to the number of components $L$ in the architecture, \ie, $M\small{=}L$.
To investigate the effect of $M$, we conduct experiments with a more different search distance $M \in \{1,3,5,10,20\}$ in MobileNet-like search space ($L=20$ in this space).

\begin{figure*}[t]
\centering
\subfloat[Ablation on \#candidate subspaces $K$.]{\includegraphics[width=0.9\columnwidth]{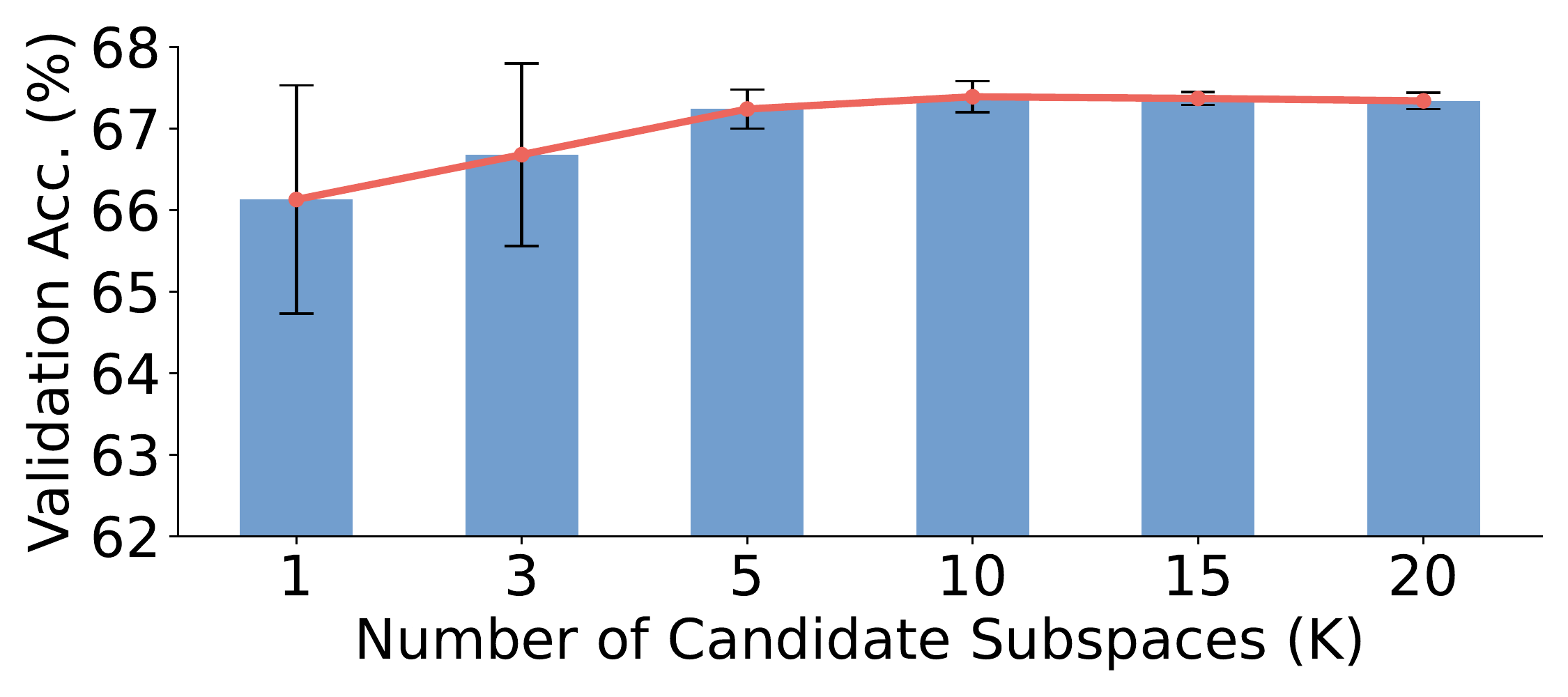}%
\label{fig:comparisons_hyperparameter_K}}
\hfil
\subfloat[Ablation on the local search distance $M$.]{\includegraphics[width=0.9\columnwidth]{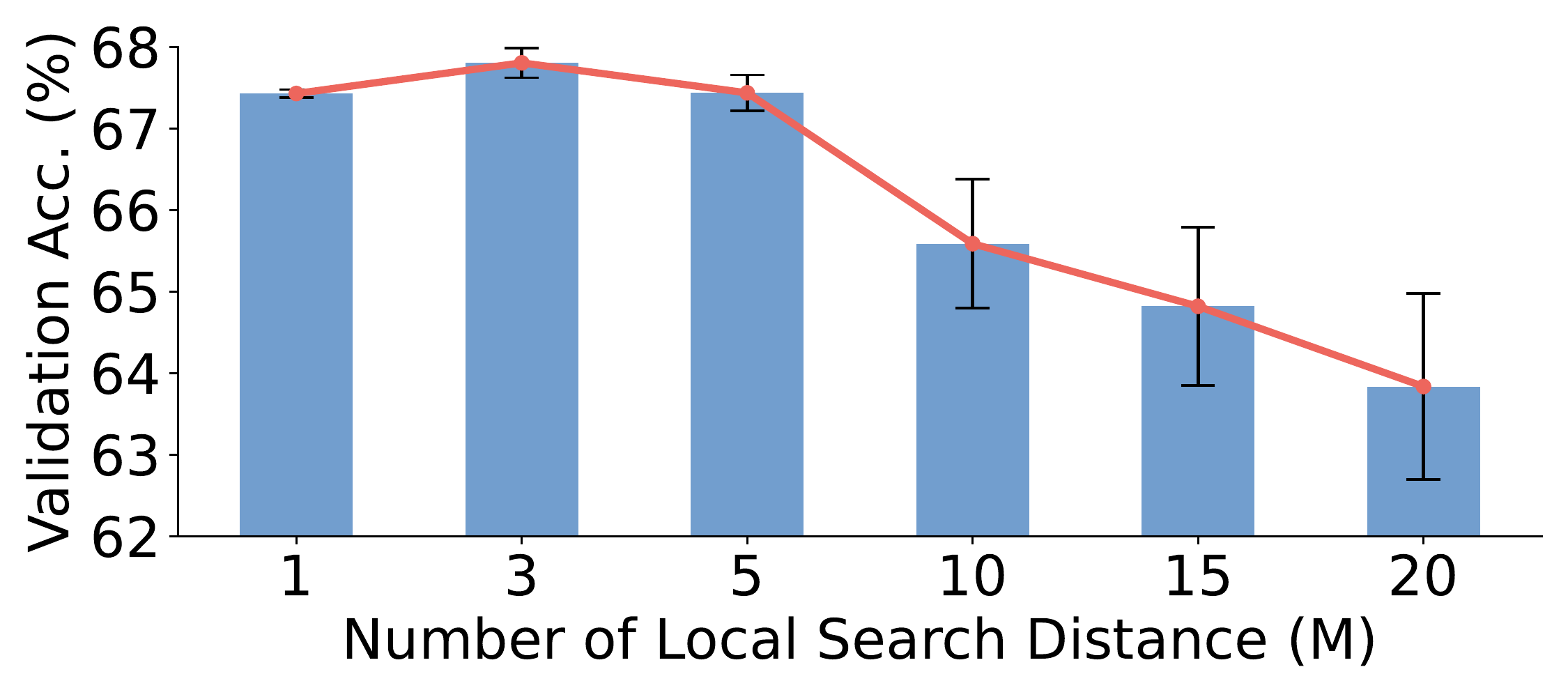}%
\label{fig:comparisons_hyperparameter_M}}
\caption{Comparisons of search performance with different candidate subspaces (a) and different local search distances (b).}
\label{fig:comparisons_hyperparameter}
\end{figure*}

In Figure~\ref{fig:comparisons_hyperparameter}(b), \sexyname achieves the best validation accuracy when $M\small{=}3$ and the worst validation accuracy when $M\small{=}20$. When $M$ is too small (\eg, $M\small{=}1$), it is easy to fall into the local optimum and hard to find better architectures in the subspace, resulting in poor search results.
When $M$ becomes larger (\eg, $M\small{>}3$), the large search space lowers the search efficiency and makes it difficult to find good architectures.
In this case, the search performance of larger search space drops greatly compared with that of small search space (\eg, only 63.84\% when $M\small{=}20$).
Thus, we set the search distance $M$ to 3 in the MobileNet-like search space.

\section{More Discussions of \sexyname}

\subsection{More Discussions on Computational Complexity}

The computational/time cost of \sexyname consists of two parts, \ie, the performance estimation cost and policy learning cost. 1) Cost of Architecture Performance Estimation. To assess the candidate architectures, we train a supernet on 10\% of the original dataset's classes, as suggested by~\cite{wu2019fbnet}. We then evaluate the validation accuracy of sub-networks on a subset of 1000 validation images. The supernet is trained with a progressive shrinking strategy~\cite{Cai2020Once} over 90 epochs. Additionally, to compute the performance improvement $R(\beta|\alpha)$ with the validation accuracy, we train a predictor to estimate validation accuracy. Together, training the supernet and the predictor requires approximately 0.75 GPU days on an NVIDIA V100 GPU. 2) Cost of Policy Learning. In Mobilenet-like search space, the controller, composed of a two-layer Graph Neural Network (GNN) and an LSTM, is trained over 10k iterations. Each iteration includes forward and backward propagations, consuming 2.73M FLOPs per iteration, totaling 27.3G FLOPs. The overall training time for the controller is about 1 GPU hour (0.04 GPU days).

\subsection{More Discussions on Advanced Subspace Search Strategies}

\sexyname facilitates extensive exploration beyond the initially identified dominant subspace through two key strategies:
1) Policy Learning for Diversity: In Eqn.~(\ref{eq:multiple_anchor_objective}), we have introduced a unique policy update strategy that prioritizes performance improvement over absolute performance. This method helps avoid local optima by encouraging the exploration of new subspaces whenever the incremental performance improvement is negligible.
This is because if the policy is in a local optimum, the performance improvement would be around zero. 
In this case, \sexyname would be encouraged to explore other subspaces in the subsequent iterations, which boosts the diversity of the explored search subspaces.
2) Dynamic Subspace Graph Update: We dynamically update the subspace graph based on the performance of newly discovered architectures. When an architecture is identified that performs better than the current ones, the subspace graph is updated to include a new subspace centered around this superior architecture. 
By only updating the subspace with a clear performance improvement, we maintain the integrity of our search results and ensure that the subspaces we identify are always promising and accurate regions of the search space.

\subsection{Mutual Influence of Global and Local Search}

In this section, we depict the mutual influence, relationship and contributions of our global and local search strategies.

1) Mutual Influence and Relationship: The global search seeks to identify a dominative subspace from a set of candidate subspaces, mining a small and effective search space for further exploration. Once a dominative subspace is determined, the local search operates within this subspace to find effective architectures. The mutual influence is evident as the results of the local search feed back into the global search. In this sense, the relationship between local and global search is complementary. When a better architecture is found through local search, we update the subspace graph with this architecture, which in turn influences subsequent global searches. This creates a feedback loop where the global search helps to guide the local search to promising regions, and the local search helps to improve the subspaces for the next global searches.

2) Contributions of Global and Local Search: By identifying small and effective subspaces, the global search significantly reduces the complexity of the search problem. This contribution is crucial in the early stages of the search process, where the need for a broad overview and quick narrowing down of the search space is imperative. On the other hand, the local search shines in its capacity for detailed exploration within the confined areas identified by the global search. Its contribution is most noticeable in meticulously exploring the narrowed-down space to find effective architectures.

\subsection{More Discussions on Initial Subspaces}

The search performance of our \sexyname depends on the initial conditions, particularly the number of initial subspaces $K$. However, our \sexyname does not need a great number of initial subspaces from three aspects:

Our proposed strategy significantly increases the diversity of the explored subspace. Our \sexyname incorporates advanced exploration strategies specifically designed to navigate through large search spaces effectively. Specifically, as detailed in Eqn.~(\ref{eq:multiple_anchor_objective}), we employ a policy optimization scheme that prioritizes performance improvement over the absolute performance of the resultant architecture.
This strategy targets subspaces with the highest potential to yield superior architectures, rather than those that merely present the best-found architecture at the moment. Our empirical analysis (refer to Figure 8(a)) demonstrates that our \sexyname is capable of exploring a more diverse range of subspaces in subsequent search phases. This significantly diminishes the risk of overfitting to initially sampled subspaces.

Transferring from previously searched architectures improves the initialization of candidate subspaces. In addition to employing a random initialization strategy for candidate subspaces, our \sexyname enhances search performance by utilizing well-designed or previously searched architectures. This is evident from the results shown in Tables~\ref{tab:nasbench_search} and~\ref{tab:imagenet}, where initializing with promising architectures leads to improved outcomes. By integrating prior knowledge of effective architectural patterns, we establish a robust mechanism for initializing candidate subspaces. This not only reduces our dependence on a large number of initial subspaces to comprehensively explore the search space but also allows us to focus our efforts on the neighborhood of promising subspaces. Such a targeted search approach effectively addresses the challenges of large search spaces and variability due to initial conditions, thereby enhancing the effectiveness of our algorithm.

Empirical studies have confirmed that $K\small{=}10$ is sufficient for exploring a large search space effectively. We have conducted ablations to investigate the effect of the number of candidate subspaces in MobileNet-like search space. It is crucial to highlight that this search space is exceptionally vast, with its size reaching up to $10^{19}$, significantly surpassing the dimensions of other adopted search spaces such as NAS-Bench-201, which has a size of $10^{5}$.
Our empirical findings reveal that setting $K\small{=}10$ is sufficient to achieve promising search performance, demonstrating that even in extensive search spaces, a modest value of $K$ can yield satisfactory results. Interestingly, when $K\small{>}10$, the performance stabilizes and exhibits negligible variations, indicating a plateau in performance gains. This suggests that the necessity for a substantial value of $K$ to ensure optimal performance as highlighted in the initial query, may not be as critical as presumed in large search spaces.

\subsection{More Discussions on Convergence Property}

\sexyname employs a dynamic subspace update strategy, a pivotal feature designed to iteratively refine the search subspace, thereby improving the search's efficacy and efficiency. This strategy is governed by a key rule: only those subspaces that yield superior performance architectures are considered for replacing and updating existing ones. This targeted updating mechanism ensures that the search is consistently steered towards more promising areas within the architecture space. Given a sufficiently large number of iterations, we anticipate that \sexyname will converge as it increasingly focuses the search on subspaces that demonstrate consistent improvements over previous iterations. This iterative refinement process, driven by a performance optimization goal, is indicative of a convergence towards an optimal or near-optimal solution in the architectural space.

While our manuscript does not provide traditional mathematical proofs of convergence and optimality, the design and operational logic of the algorithm, particularly the dynamic subspace update strategy, offer a conceptual basis for expecting properties of convergence and optimality.
Additionally, the experimental results presented in Tables II and IV validate the effectiveness of \sexyname. Together, the analytical insights and empirical evidence presented address the concerns about the theoretical underpinnings of our algorithm.

\section{Conclusion}

In this paper, we proposed a Neural Architecture Search method via Dominative Subspace Mining (\sexyname), which focuses on automatically mining small and effective subspaces and conducts a search in them.
Specifically, we first perform a global search for a dominative subspace from the candidate subspaces.
Then, we perform a local search for effective architectures in the globally searched subspace instead of the original large one.
Finally, we update the candidate subspace with the locally searched architecture.
Moreover, \sexyname further enhance search performance by taking well-designed/searched architectures as the prior knowledge.
Extensive experiments on several benchmark search spaces demonstrate the superiority of our method over the considered methods.

\bibliographystyle{IEEEtran}
\bibliography{IEEEabrv,egbib}

\begin{IEEEbiography}[{\includegraphics[width=1in,height=1.25in,clip,keepaspectratio]{{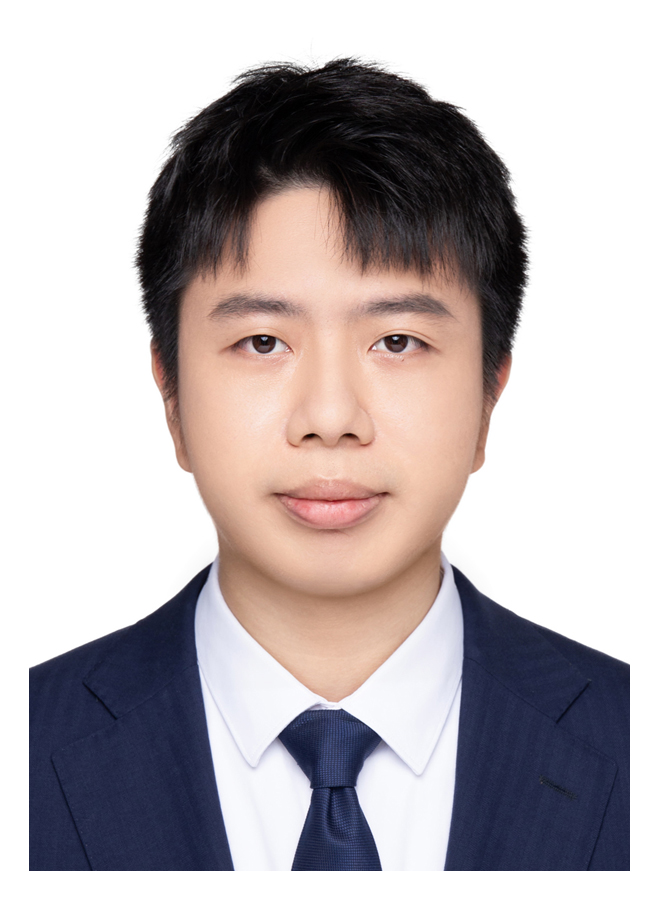}}}]{Yaofo Chen}
received the B.E. degree in Software Engineering from South China University of Technology, China, in 2018. He is currently pursuing the Ph.D. degree in the School of Software Engineering, South China University of Technology, China. His research interests include Neural Architecture Search and Computer Vision.
\end{IEEEbiography}

\begin{IEEEbiography}[{\includegraphics[width=1in,height=1.25in,clip,keepaspectratio]{{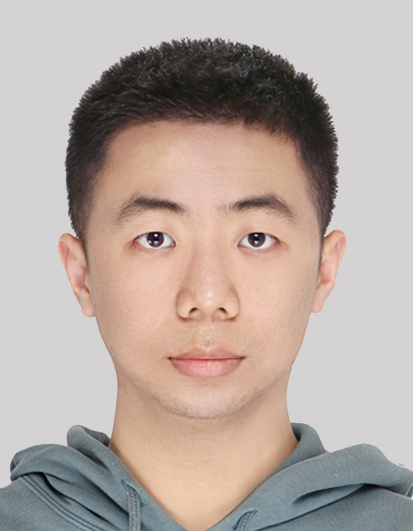}}}]{Yong Guo}
is currently a postdoctoral researcher in Max Planck Institute for Informatics. He received his bachelor's degree from South China University of Technology in 2016 and his Ph.D. degree from the same university in 2021. His research interests include deep learning and computer vision.
\end{IEEEbiography}

\begin{IEEEbiography}[{\includegraphics[width=1in,height=1.25in,clip,keepaspectratio]{{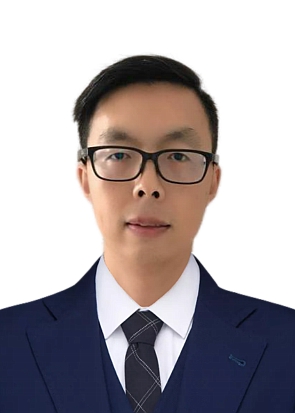}}}]{Daihai Liao}
is currently a Senior Artificial Intelligence Engineer with Changsha Hisense Intelligent System Research Institute Co., Ltd and has applied for 5 invention patents.
He received his Master degree in Control Science and Engineering from Central South University, China.
His research interests include object detection, object tracking and semantic segmentation.
\end{IEEEbiography}

\begin{IEEEbiography}[{\includegraphics[width=1in,height=1.25in,clip,keepaspectratio]{{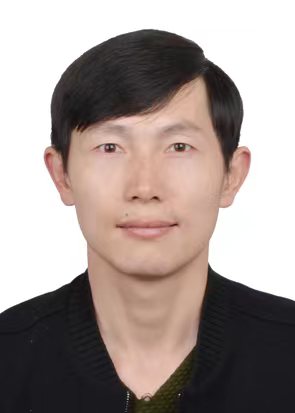}}}]{Fanbing Lv}
is currently a Deputy Chief Engineer, and Senior Engineer of Big Data with Changsha Hisense Intelligent System Research Institute Co., Ltd and has applied for 20 invention patents.
He was selected into Guiyang Big Data Hundred Talents Program and High-level Talent Green Card (Class C) and won the "First Prize" in the Fourth Guizhou Innovative Product Design Competition
\end{IEEEbiography}

\begin{IEEEbiography}[{\includegraphics[width=1in,height=1.25in,clip,keepaspectratio]{{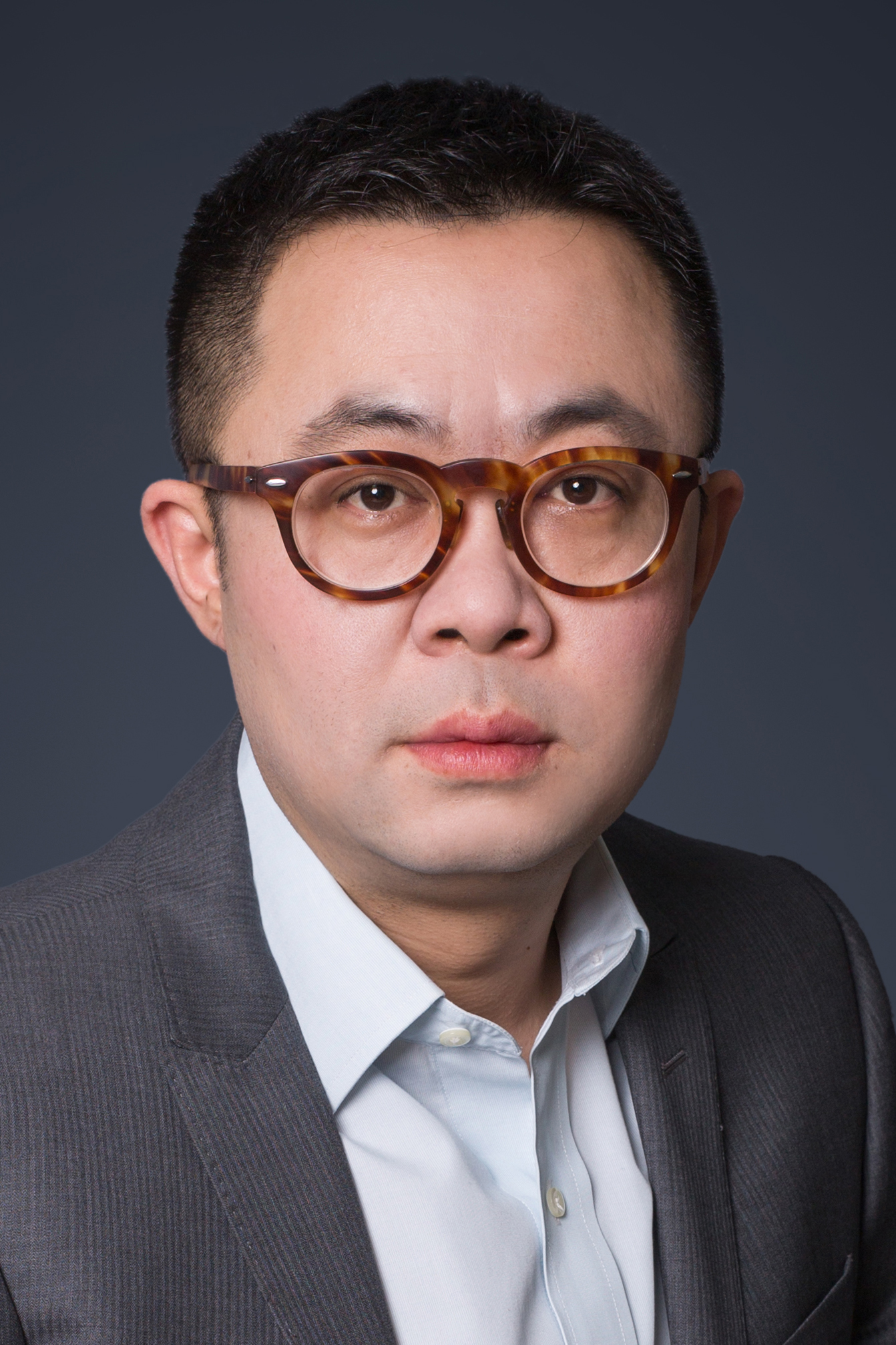}}}]{Hengjie Song}
is currently a Professor with the School of Software Engineering, South China University of Technology.
He has published several high quality articles on the best journals and conference proceedings,
including the \textit{ACM Transactions on the Web}, the IEEE CIMs, the IEEE \textsc{Transactions on Fuzzy Systems}, \textit{Neural Networks} (Elsevier), AAAI, and ICDM.
His research interests include artificial intelligence and the applications of AI in commercial search engines.
\end{IEEEbiography}

\begin{IEEEbiography}[{\includegraphics[width=1in,height=1.25in,clip,keepaspectratio]{{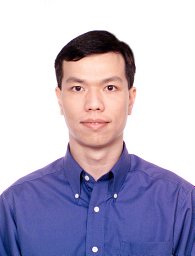}}}]{James Tin-Yau Kwok}
(Fellow, IEEE) received the Ph.D. degree in computer science from The Hong Kong University of Science and Technology, Hong Kong, in 1996. He is currently a Professor with the Department of Computer Science and Engineering, The Hong Kong University of Science and Technology. His current research interests include kernel methods, machine learning, pattern recognition, and artificial neural networks. He received the IEEE Outstanding Paper Award in 2004 and the Second Class Award in Natural Sciences from the Ministry of Education, China, in 2008. He has been a Program Co-Chair for a number of international conferences and served as an Associate Editor for the IEEE TRANS-ACTIONS ON NEURAL NETWORKS AND LEARNINGSYSTEMS from 2006 to 2012. He is currently an Associate Editor of Neurocomputing.
\end{IEEEbiography}

\begin{IEEEbiography}[{\includegraphics[width=1in,height=1.25in,clip,keepaspectratio]{{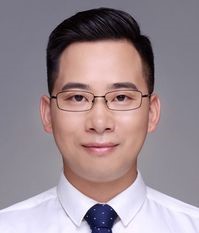}}}]{Mingkui Tan}
is currently a professor with the School of Software Engineering at South China University of Technology. He received his Bachelor Degree in Environmental Science and Engineering in 2006 and Master degree in Control Science and Engineering in 2009, both from Hunan University in Changsha, China. He received his Ph.D. degree in Computer Science from Nanyang Technological University, Singapore, in 2014. From 2014-2016, he worked as a Senior Research Associate on computer vision in the School of Computer Science, University of Adelaide, Australia. His research interests include machine learning, sparse analysis, deep learning, and large-scale optimization.
\end{IEEEbiography}

\end{document}